\documentclass[lettersize,journal]{IEEEtran}
\usepackage{amsmath,amsfonts}
\usepackage{algorithmic}
\usepackage{algorithm}
\usepackage{array}
\usepackage{booktabs}
\usepackage{textcomp}
\usepackage{stfloats}
\usepackage{url}
\usepackage{verbatim}
\usepackage{graphicx}
\usepackage{cite}
\usepackage{hyperref}
\usepackage{multirow}
\hypersetup{breaklinks=true}

\usepackage{subcaption}
\usepackage{tikz}
\usetikzlibrary{positioning,shapes,arrows,calc}
\usepackage{ragged2e}

\usepackage{xcolor}

\DeclareMathOperator*{\argmin}{arg\,min}

\begin{document}

\title{Deep Self-Supervised Disturbance Mapping with the OPERA Sentinel-1 Radiometric Terrain Corrected SAR Backscatter Product}

\author{Harris Hardiman-Mostow, Charles Marshak, Alexander L. Handwerger
\thanks{This work has been submitted to the IEEE for possible publication. Copyright may be transferred without notice, after which this version may no longer be accessible. \textit{(Corresponding author: Harris Hardiman-Mostow.)}

Harris Hardiman-Mostow is with the Department of Mathematics, University of California - Los Angeles, Los Angeles CA 90095 USA (email: hhm@math.ucla.edu).
He is also currently a research affiliate with the Jet Propulsion Laboratory, California Institute of Technology, Pasadena CA 91109 USA.

Charles Marshak is with the Jet Propulsion Laboratory, California Institute of Technology, Pasadena CA 91109 USA (email: charlie.z.marshak@jpl.nasa.gov).

Alexander L. Handwerger is with the Jet Propulsion Laboratory, California Institute of Technology, and Joint Institute for Regional Earth System Science and Engineering, University of California - Los Angeles (email: alexander.handwerger@jpl.nasa.gov).

}
}

\markboth{Submitted to IEEE Journal of Selected Topics in Applied Earth Observations and Remote Sensing}{Hardiman-Mostow \MakeLowercase{\textit{et al..}}: Deep Self-Supervised  Disturbance Mapping with Sentinel-1 OPERA RTC Synthetic Aperture Radar}


\maketitle

\begin{abstract}

Mapping land surface disturbances supports disaster response, resource and ecosystem management, and climate adaptation efforts. 
Synthetic aperture radar (SAR) is an invaluable tool for disturbance mapping, providing consistent time-series images of the ground regardless of weather or illumination conditions. 
Despite SAR's potential for disturbance mapping, processing SAR data to an analysis-ready format requires expertise and significant compute resources, particularly for large-scale global analysis.
In October 2023, NASA's Observational Products for End-Users from Remote Sensing Analysis (OPERA) project released the near-global Radiometric Terrain Corrected SAR backscatter from Sentinel-1 (RTC-S1) dataset, providing publicly available, analysis-ready SAR imagery.
In this work, we utilize this new dataset to systematically analyze land surface disturbances.
As labeling SAR data is often prohibitively time-consuming, we train a self-supervised vision transformer - which requires no labels to train - on OPERA RTC-S1 data to estimate a per-pixel distribution from the set of baseline imagery and assess disturbances when there is significant deviation from the modeled distribution.
To test our model's capability and generality, we evaluate three different natural disasters - which represent high-intensity, abrupt disturbances - from three different regions of the world.
Across events, our approach yields high quality delineations: $F_1$ scores exceeding $0.6$ and Areas Under the Precision-Recall Curve exceeding $0.65$, consistently outperforming existing SAR disturbance methods. 
Our findings suggest that a self-supervised vision transformer is well-suited for global disturbance mapping and can be a valuable tool for operational, near-global disturbance monitoring, particularly when labeled data does not exist.

\end{abstract}


\section{Introduction}
\label{sec:intro}
\IEEEPARstart{T}{he} Earth's surface is continually changing due to a combination of natural \cite{marlon2012long,larsen2012landslide,peltier2004global} and human-induced factors \cite{curtis2018classifying,hansen2013high}.
While some changes occur gradually over time (e.g., mountain formation, glacial isostatic adjustment), others are abrupt and can have immediate impacts on the environment (e.g., landslides, wildfires, flooding, deforestation, mining). 
Satellite remote sensing can detect these surface changes and disturbances over large areas and on various time scales to address critical environmental and societal challenges \cite{rose2015ten,knight2013impacts}.
Through disturbance delineation, we enhance our capability to monitor vulnerable or protected habitats \cite{rose2015ten}, map the impacts of human urbanization and development \cite{avtar2020assessing}, 
track population migration patterns \cite{meyer1992human},
identify tropical forest loss \cite{hansen2013high, reiche_radd_2021_forest}, 
and respond to natural hazards such as earthquakes \cite{stephenson2021deep}, floods \cite{pengflood, sen1_floods}, landslides \cite{handwerger2022generating,amatya2021landslide}, and fires \cite{ban2020near, rignot1993change}. 

Among remote sensing techniques, synthetic aperture radar (SAR) is an invaluable modality for disturbance mapping, with the capability to rapidly image the ground  regardless of cloud cover or illumination conditions. 
While optical imagery provides information about \emph{spectral} changes, SAR provides information about \emph{structural} changes that are captured via backscatter and phase. Applications include centimeter-scale motion of the Earth's surface \cite{rosen2000synthetic,biggs2020satellite}, crop mapping \cite{han2023spatio}, structural damages after earthquakes \cite{stephenson2021deep}, and landslide  \cite{jung2020evaluation, jung2017damage}, and wildfire delineation \cite{rignot1993change, ban2020near}.

Despite the proven capability of SAR for disturbance mapping, analysis-ready SAR data has - until recently - not been immediately available, requiring expert processing to produce
such datasets over large areas and timescales.
However, the release of the near-global Radiometric Terrain Corrected SAR backscatter from the C-band Sentinel-1 (RTC-S1) dataset by NASA's Observational Products for End-Users from Remote Sensing Analysis (OPERA) project in October 2023 has effectively removed this barrier for Sentinel-1. 
OPERA RTC-S1 provides publicly available dual-polarization\footnote{The two polarizations are co-polarized and cross-polarized, also known as vertical transmit and vertical receive (VV) and vertical transmit and horizontal receive (VH).} SAR backscatter imagery in an analysis-ready format \cite{opera_validation_plan}.
Moreover, RTC-S1 mitigates the effects of terrain and SAR acquisition geometry, such as layover and shadow, that can obfuscate surface changes observed by SAR \cite{small_rtc, shiroma2023opera}.
These three qualities - analysis-ready, near-global, and terrain correction - makes OPERA RTC-S1 an ideal dataset for systematic and thematic disturbance delineation.

The aim of this work is to train a \emph{self-supervised}, transformer-based model capable of probabilistically delineating \emph{generic} disturbances in OPERA RTC-S1 time-series.
Here, \emph{generic} disturbances refers to any changes that deviate from the set of baseline imagery used to establish a range of nominal observations of the land surface.

\textit{Supervised} machine learning techniques rely on the existence of a large, labeled training set.
However, labeling such datasets for disturbance mapping - particularly global-scale SAR time series spanning months or years - is impractical: it requires expertise of SAR physical mechanisms \cite{woodhouse2017introduction, oliver_quegan_2004understanding_sar} as well as knowledge of the land cover being monitored to determine what signals are phenological and which are true disturbances. 
Despite this, many previous works on SAR disturbance mapping are supervised \cite{ban2020near, venkataramani2023harmonizing, nava2021improving, nava2024sentinel}, requiring carefully curated, labeled training sets. Not only is this difficult to scale to larger datasets, but the models trained on these specialized datasets lack the generality to be a broadly applicable, generic disturbance model.
While there are numerous optical disturbance datasets \cite{Chen2020_levir_CD, whu_cd, huggingface_hls_burn_scars, s12_floods} and corresponding supervised models (see e.g. \cite{chen2024changemamba}), training a SAR model on optically-derived labels presents significant challenges as some disturbances visible to an optical sensor will be invisible to SAR and vice versa.
Additionally, with forthcoming NASA ISRO SAR (NISAR) data, where the creation of a large labeled corpus from the new L-band imagery is not possible, the demonstration of a label-free approach for SAR-based applications is essential.


This motivates \textit{self-supervised} learning for SAR-based disturbance mapping. Self-supervision remedies the need for labels by generating a supervised signal from the data itself \cite{he2022masked, manas2021seasonal, chen2020simple}. One previous performant self-supervised work on SAR damage mapping was by Stephenson et al. \cite{stephenson2021deep}, which utilized a recurrent neural network (RNN) \cite{rumelhart1986learning} on interferometric SAR coherence to map earthquake damage extents (conversely, this work considers SAR backscatter). RNNs are neural networks designed to handle temporal data, such as language processing tasks \cite{sutskever2014sequence}. However, more recently, RNNs have largely been supplanted by transformers \cite{vaswani2017attention}, which are now state-of-the-art in a variety of tasks, including language \cite{NEURIPS2020_1457c0d6} and vision \cite{dosovitskiy2020image}, as well as remote sensing tasks \cite{cong2022satmae,tseng2023lightweight}. Moreover, transformers avoid exploding and vanishing gradients, which tend to plague RNNs. Hence, this work will train a transformer in a self-supervised fashion. We also improve on \cite{stephenson2021deep} by training and deploying a single model on multiple types of damage events in three different parts of the world, whereas \cite{stephenson2021deep} trained separate models for each location, and only focused on earthquake damage extents.

\subsection{Our Contributions}
\label{sec:contributions}

In this work, we train a self-supervised vision transformer \cite{dosovitskiy2020image} on near-global OPERA RTC-S1 imagery and demonstrate superior damage mapping performance across environments and disaster events compared to previous works.
Our transformer learns to estimate the per-pixel distribution from baseline of OPERA RTC-S1 imagery. 
This estimated distribution can be compared to a new acquisition by computing the Mahalanobis distance, yielding a disturbance metric that can be used to delineate disturbance.
The metric has a statistical interpretation as the number of standard deviations (SD) from the modeled per-pixel mean.
More simply put, the metric quantifies how likely a pixel is disturbed, with larger values indicating greater likelihood of disturbance.
The vision transformer is trained in a self-supervised manner, meaning no labels or human annotation are required; the only input is the OPERA RTC-S1 imagery. 
We evaluate our disturbance metric on external validation data from three recent natural disasters: 1) a landslide in Papua New Guinea, 2) a series of fires in Chile, and 3) flooding in Bangladesh. 
We use the same model and metric for every event. 
The validation data is derived from publicly-available damage maps \cite{UNOSAT, copernicus_rapid_mapping}; much of which is made from Very High Resolution (VHR) optically-derived inputs.
These natural disasters represent high-intensity disturbances where we expect significant alignment between optical- and SAR-based disturbance delineations.
The validation data used for this work, including the disturbance delineations and data provenance, are found in the GitHub repository \cite{opera_dist_s1_events}.
We demonstrate that the proposed vision transformer-based disturbance model has systematically better performance compared to an RNN-based model \cite{stephenson2021deep} and the classical log ratio method for SAR change mapping \cite{rignot1993change}. Code for our data\footnote{\url{https://github.com/OPERA-Cal-Val/dist-s1-events}} and experiments\footnote{\url{https://github.com/opera-adt/distmetrics}} are available via GitHub.

\subsection{Organization of Paper}

The remainder of this work is organized as follows. In Section \ref{sec:background}, we review previous works on disturbance mapping with SAR, transformers, and self-supervision. In Section \ref{sec:methodology}, we detail the mathematical formulation of our model, our disturbance metric for computing delineations, and the model architectures and training. Sections \ref{sec:data} and \ref{sec:disaster_data} describe the OPERA RTC-S1 data (and external validation data) we use to train and evaluate our model, respectively. Section \ref{sec:results} presents our disturbance mapping results on three natural disasters, where our transformer-based model consistently outperforms previous works (RNN and Log Ratio). We also present ablation experiments that justify some of our model choices. 
Although our evaluation is limited to three disasters, the ability to accurately delineate different categories of disturbances in varied environments using a single vision transformer provides some encouraging evidence that such an approach may be applicable to monitor disturbance at larger scales.
To this end, we will discuss future work to further evaluate this model in Section \ref{sec:discussion} and \ref{sec:conclusion}.

\section{Background}\label{sec:background}

\begin{figure*}[t]
    \centering
    \includegraphics[width=.9\textwidth]{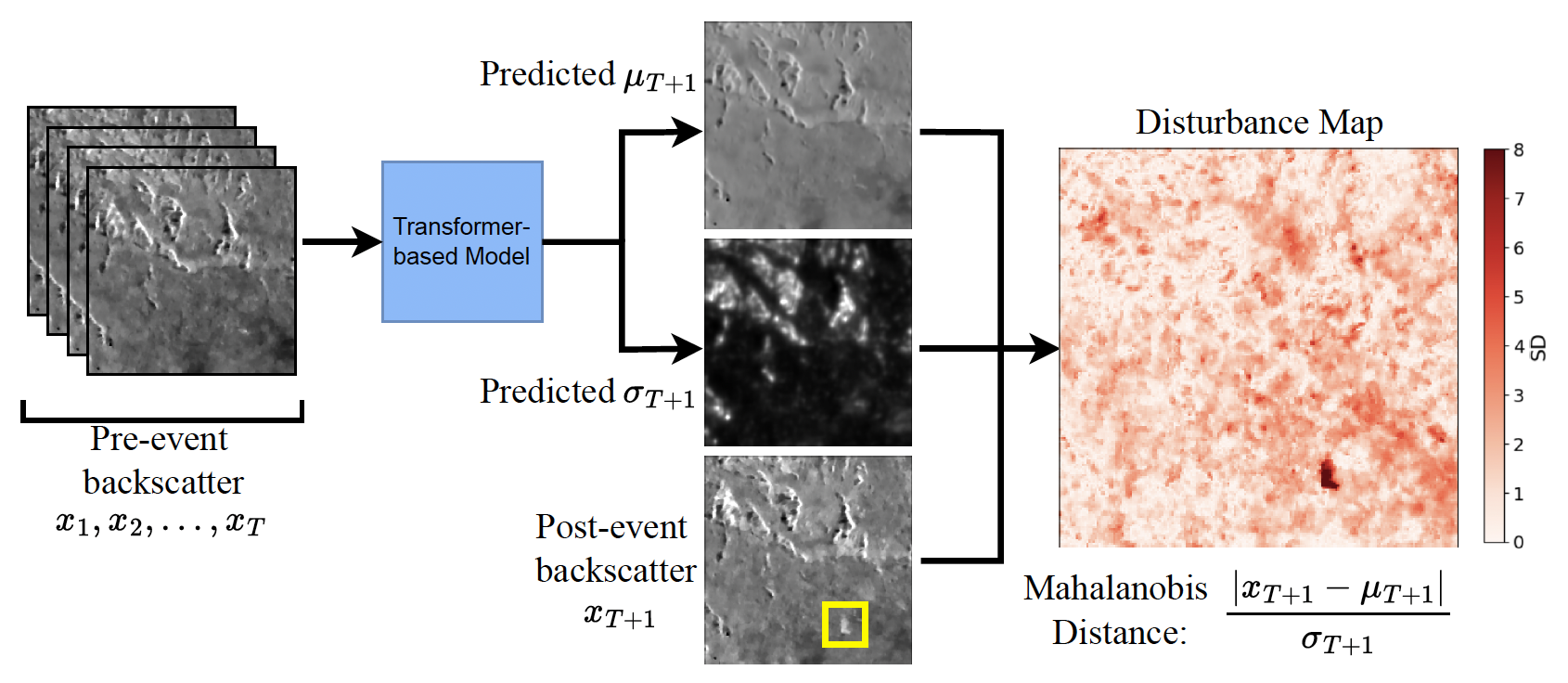}
    \caption{Visualization of the disturbance metric and how it is computed for single channel (VV) time-series. This example shows data from a landslide that occurred in Enga Province, Papua New Guinea on May 24, 2024. We use a sequence of baseline imagery $x_1, x_2, ... x_T$ (left) from before the event to  predict a distribution for each pixel in the image with mean $\mu_{T+1}$ and standard deviation $\sigma_{T+1}$ (center right). These parameters are used to quantify disturbance in the observed image $x_{T+1}$ (center right, bottom) to compute one-dimensional transformer metric via the Mahalanobis distance (right), where higher values (measured in standard deviations or SD) indicate a higher likelihood of disturbance. Our methodology is able to precisely delineate the landslide disturbance extents (bottom right corner of the Disturbance Map).}
    \label{fig:schematic}
\end{figure*}

\subsection{Disturbance Mapping}
\label{sec:disturbance_mapping}

Disturbance mapping is the process of identifying pixels in space and time that undergo changes (i.e. disturbances) within a coregistered time-series of remotely-sensed imagery.
Disturbance mapping is sometimes referred to as ``change detection'' in the remote-sensing literature \cite{canty2019image_change_detection, zhu2017deep}.
While disturbance mapping encompasses many types of changes, our work will focus specifically on abrupt, sudden disturbances (i.e., damage) caused by natural disasters.
Natural disasters are a key application area for disturbance models and simplify validation by providing a clearly defined point in a time series to map disturbances.
Hence, we will use ``change'', ``damage",  and ``disturbance'' interchangeably. 
In this work, we are focused on \emph{generic} Sentinel-1 disturbance delineation - that is, modeled statistical deviation of the current acquisition from a set of baseline images.
Generally, what disturbances can be mapped depends on the temporal sampling of inputs, the spatial resolution, the type of change being monitored, the modality of the sensor, and the data used to construct the baseline.
Sentinel-1 data, and hence OPERA RTC-S1 data, currently has 12-day sampling frequency\footnote{Sentinel-1 A/B constellation had a 6 day repeat pass frequency. Since Sentinel-1 B's decommission in December 2021, Sentinel-1 A is the only satellite collecting data and has 12 day repeat pass frequency. When Sentinel-1 C is launched in December 2024, the constellation will again have 6 day temporal sampling.}, though not all Sentinel-1 passes result in new data \cite{asf_sentinel1_acquisition_maps}.
While this means some disturbances may not be resolved by Sentinel-1 (a flash flood can occur and resolve between samples), our approach leverages the available data's strengths and is applicable to \emph{any} time series generated by Sentinel-1.  

A similar view of generic disturbance is found in OPERA's near-global \textit{optical} Land Surface Disturbance from Harmonized Landsat Sentinel-2 (DIST-HLS) generic disturbance algorithm \cite{opera_validation_plan, dist_hls_atbd}.
The DIST-HLS algorithm estimates the per-pixel mean and standard deviation using sample statistics from multiple monthly composites and measures deviation from the baseline via the Mahalanobis distance \cite{dist_hls_atbd}.
In this work, we develop an algorithm that provides a complementary measure of generic disturbances using SAR, with the future goal of deploying it at the near-global scale.

Currently, there are two operational\footnote{Here, operational here means that each new SAR acquisition triggers the generation of a new disturbance map.} SAR disturbance products, which track tropical forest loss: Advanced Land Observing Satellite (ALOS) forest alerts \cite{koyama_forest_alos_maps, reiche2021forest} and RAdar for Detecting Deforestation (RADD) \cite{reiche_radd_2021_forest, reiche_radd_algorithm}.
Both rely on estimating backscatter conditioned on ``forest" and ``non-forest" distributions and delineating disturbance when there is evidence forest has been removed based on these modeled distributions.
Our approach aims to delineate a broader class of disturbances, across terrains and disturbance events. To this end, we utilize a vision transformer trained on near-global RTC-S1 data to determine the range of nominal backscatter values in varied environments and land cover types, not just dense tropical forest.

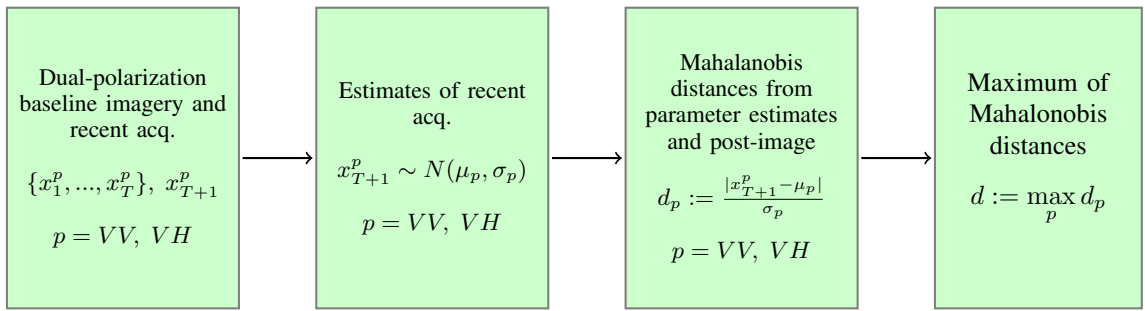
\begin{figure*}[t]
    \centering
    \begin{tikzpicture}[
    box/.style={rectangle, draw=black!50, fill=green!20, thick, minimum width=2.75cm, minimum height=4.cm, text centered},
    arrow/.style={->, thick, shorten >=1pt, shorten <=1pt}
    ]
    
     \node[box] (box1){\parbox{2.85cm}{\centering \small Dual-polarization baseline imagery and recent acq. \\ \; \\ $\{x_1^p, ..., x^p_T\}, \; x_{T+1}^p$\\ \; \\ $p = VV,\; VH$ }};
    \node[box, right=of box1] (box2) {\parbox{2.85cm}{\centering \small Estimates of recent acq. \\ \; \\$x^{p}_{T+1} \sim N(\mu_{p},\sigma_{p})$\\ \; \\ $p = VV,\; VH$}};
    \node[box, right=of box2] (box3) {\parbox{2.85cm}{\centering \small Mahalanobis distances from parameter estimates and post-image \\ \; \\  $d_{p}:= \frac{|x^p_{T+1} - \mu_{p}|}{ \sigma_{p}}$ \\ \; \\ $p = VV,\; VH$}};
    \node[box, right=of box3] (box4) {\parbox{2.5cm}{\centering Maximum of  Mahalonobis distances  \begin{equation*}d:= \max_{p} d_{p} \end{equation*}}};

    \draw[arrow] (box1) -- (box2);
    \draw[arrow] (box2) -- (box3);
    \draw[arrow] (box3) -- (box4);
\end{tikzpicture}
    \caption{A flow chart illustrating the transformer metric proposed in this paper. The metric is computed for dual polarization images (see Fig. \ref{fig:schematic} for 1-dimensional visualization). The first arrow is where the deep-learning model is utilized to estimate the per-pixel distribution of the $x_{T+1}$.}
    \label{fig:metric-flow}
\end{figure*}

In order to contextualize our approach, it is helpful to define the notion of a ``disturbance metric.'' 
A disturbance metric compares a baseline set of images to a recent image acquisition to determine the per-pixel likelihood of disturbance \cite{deledalle2012compare}.
Larger values of the metric typically indicate greater likelihood of disturbance.
Using a disturbance metric, delineation of disturbances (that is, classifying each pixel as ``disturbed" or ``undisturbed") is simplified to thresholding the metric.
For example, the classical log-ratio method \cite{rignot1993change} defines a disturbance metric that identifies changes based on the decibel (dB) deviation from the baseline image set:
\[
\ell:= \left |\log_{10} \left(\frac{I_1}{I_0}\right)\right| = \left|\log_{10}(I_1) - \log_{10}(I_0)\right|.
\]

Above, $I_1$ is the post-event cross- or co-polarization image and $I_0$ is a pre-event cross- or co-polarization image. $I_0$ is also sometimes an aggregated reference image; in our experiments, we will set $I_0$ to be the median of the available pre-event imagery. 
Although it was developed over 30 years ago, log ratio is still widely used in modern SAR damage mapping research \cite{aimaiti2022war,handwerger2022generating,jung2020evaluation,li2022augmentation,mondini2019sentinel,mondini2021landslide,peters2024detecting}.
It is well known that large negative values in $\log(I_1) - \log(I_0)$ indicate vegetation loss when we consider  cross-polarization data \cite{rignot1993change, woodhouse2017introduction}.  
Above, we consider the absolute value because we aim to delineate generic disturbances from a baseline set of imagery, which could be indicated by anomalously low or high measurements in either polarization.
In Section \ref{sec:metric}, we will define a disturbance metric for our transformer-based approach, which compares favorably to $\ell$ in our experiments (Section \ref{sec:results}).

Similarly, one can fashion likelihood tests to generate disturbance metrics \cite{deledalle2012compare, rabasar}.
Although a disturbance metric is a per-pixel measure, it is  often helpful to incorporate nearby pixel information to better contextualize disturbance e.g. through superpixels \cite{marshak_superpixels}, convolutional layers of a neural network \cite{bergamasco2019unsupervised_bruzzone_conv_net}, or Markov random fields \cite{bruzzone_diff}.
In this work, our vision transformer is designed to estimate the per-pixel distribution from the baseline set using rich spatio-temporal information extracted from the data via the neural network architecture.
We utilize the learned distributional parameters to construct a disturbance metric via the Mahalanobis distance between this estimated distribution and the most recent image.
This metric has a convenient statistical interpretation as the number of (estimated) standard deviations away from the (estimated) mean at each pixel.

The goal of our disturbance algorithm is to develop a methodology that can identify disturbance with each new Sentinel-1 acquisition, utilizing the preceding acquisitions to establish a baseline \cite{devries2020rapid, reiche2018characterizing,reiche2021forest, handwerger2022generating}.
While this process is well-suited over large areas and multiple environments, there are also change detection methods focused on bi-temporal comparisons, i.e. using a single image pair, in which an abrupt disturbance and often damages from a natural disaster can be observed \cite{corley_caleb_robinson_2024change, mou2018learning_st_features, whu_cd, Chen2020_levir_CD, bruzzone_unsupervised, jung2020evaluation,yun2015rapid}.
Some algorithms, such as the proposed NISAR CuSum algorithm \cite{nisar2018handbook, nisar2022productgeneration_atbd_cusum, sabir2024adapting_cusum}, analyze a full year of imagery and then identify the changed areas in space and time from the available data. 
Our transformer-based approach can use anywhere from 2-10 acquisitions; all of our tests used at least 4 acquisitions (representing approximately a six week period, see \ref{sec:results_flood}).

Our work utilizes OPERA RTC-S1, a near-global dual-polarization backscatter product.
Backscatter amplitude is one component of SAR remote sensing.
The interferometric phase and coherence are invaluable for damage assessment and disturbance \cite{stephenson2021deep, jung_ash_coherence, jung2020evaluation}.
While incorporating such data into our models would provide richer structure, there is not a near-global interferometric phase product.
The OPERA Coregistered Single Look Complex dataset (OPERA CSLC-S1), which contains phase information from Sentinel-1, is only available over North America.
Even this higher-level product would require significant pre-processing to incorporate into our model and such processing is beyond the current scope of this work.

\subsection{Transformers and Self-Supervised Training}

Transformers \cite{vaswani2017attention} are becoming the de-facto architecture across numerous machine tasks, producing state-of-the-art results in language modeling \cite{openai2024gpt4technicalreport, devlin2018bert}, image classification \cite{dosovitskiy2020image, he2022masked}, and semantic segmentation in imagery and video \cite{he2022masked, video_mae, kirillov2023segment, feichtenhofer2022masked}, among other areas. 
Empirically, transformers are highly generalizable and can be fine-tuned on downstream tasks with relatively small labeled datasets  \cite{he2022masked, kirillov2023segment, NEURIPS2020_1457c0d6}. 
Transformer's ability to learn new tasks quickly has given rise to ``foundation models'' \cite{yuan2021florence, lacoste2024geobench}.
These models are becoming standard in geospatial data analysis \cite{cong2022satmae, tseng2023lightweight, reed2023scale, jakubik2023foundation, clayfoundation2024model, stewart2024ssl4eo}.
These geospatial transformers too can be fine-tuned across many standard remote sensing analyses including delineation of flood extents, wildfire perimeters, and landslides \cite{jakubik2023foundation}, estimation of biomass, agricultural yield, and tree height \cite{clayfoundation2024model, tseng2023lightweight, cambrin2024tree_height}, and generation of land-use maps at the continental scale \cite{allen2023large}.
Typically, these models are trained in a self-supervised fashion, meaning that no labeled data (or very limited labeled data for fine-tuning) is required to produce results.

Self-supervised learning is a training methodology in which a supervised learning signal is created by the input data itself. 
There are numerous approaches to self-supervised training including autoencoding \cite{kingma2013auto}, residual learning/backward diffusion \cite{dncnn_denoising, ho2020denoising}, masked autoencoding \cite{he2022masked, tong2022videomae, feichtenhofer2022masked, NEURIPS2020_1457c0d6} and contrastive learning \cite{chen2020simple}. 
In remote sensing, self-supervised training is invaluable as it removes the requirement to create enough labels to train deep neural network models.
Moreover, it allows these deep models to be quickly adapted to various sensors, each of which may have unique sensitivities, varying acquisition patterns, and often large unlabeled data archives.
Indeed, self-supervised learning is becoming the standard way to train these transformer-based models for geospatial analysis  \cite{reed2023scale, tseng2023lightweight, manas2021seasonal, kuzuforest, reed2023scale, allen2023large, clayfoundation2024model, jakubik2023foundation, lacoste2024geobench}. 
Our self-supervised approach will follow the method proposed in \cite{stephenson2021deep}, where our models will predict a pixel-based distribution for an image and use a statistical measure (z-score) to map disturbances.
The model used in \cite{stephenson2021deep} is an RNN, focuses only on disturbances caused by earthquakes, and trains a new model for each earthquake-affected area studied. We will show that the \textit{transformer}-based architecture improves disturbance delineation over an RNN for the OPERA RTC-S1 data, and that a \textit{single} transformer-based architecture can be used across multiple environments and across multiple categories of disasters.

We will briefly review some key components of transformers.
At the heart of transformers is the attention mechanism, which computes an attention ``score" based on three learned matrices: the query $Q$, the key $K$, and the value $V$.
\begin{equation}
\text{Attention}(Q, K, V) = \text{softmax}\Bigl( \frac{QK^T}{\sqrt{d_k}} \Bigr) V,
\end{equation}
where $d_k$ is the dimension of $Q$ and $V$. Attention allows the model to learn how different elements of a sequence ``attend" to one another, learning complex temporal relationships without the exploding or vanishing gradient issue of RNNs. Another novelty of transformers is their positional encoding (these can be either learned or fixed - we will use learned), which is added to the input embeddings to indicate position to the model. Combined with residual connections \cite{he2016deep} and layer normalization \cite{lei2016layer}, transformers have been shown to be a powerful tool in machine learning.

\section{Methodology}
\label{sec:methodology}

\begin{figure}[!t]
    \centering
    
    \begin{subfigure}{.49\linewidth}
    \centering
    \includegraphics[width=\linewidth]{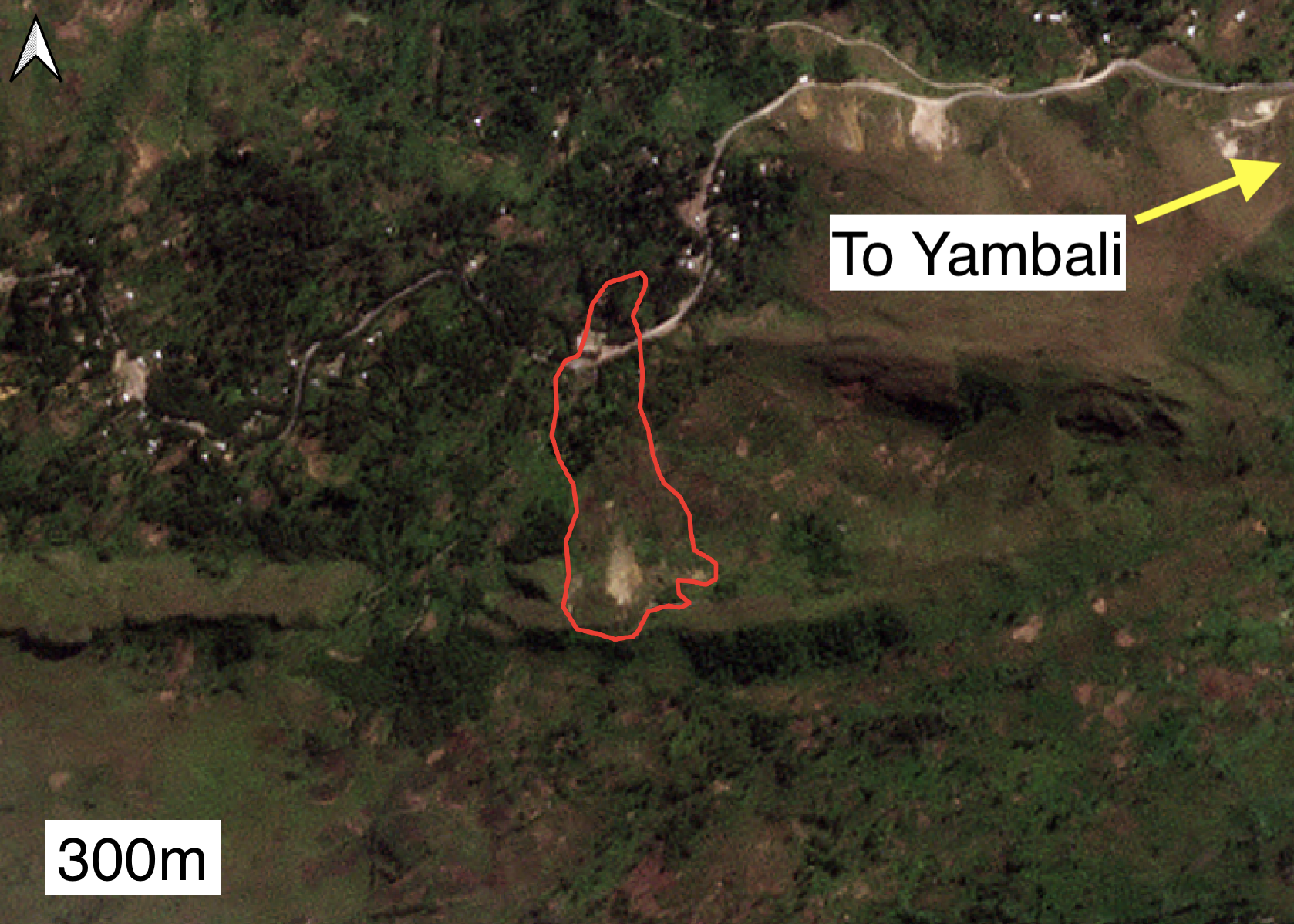}
    \caption{May 15, 2024}
    \end{subfigure}
    \hfill
    \begin{subfigure}{.49\linewidth}
    \centering
    \includegraphics[width=\linewidth]{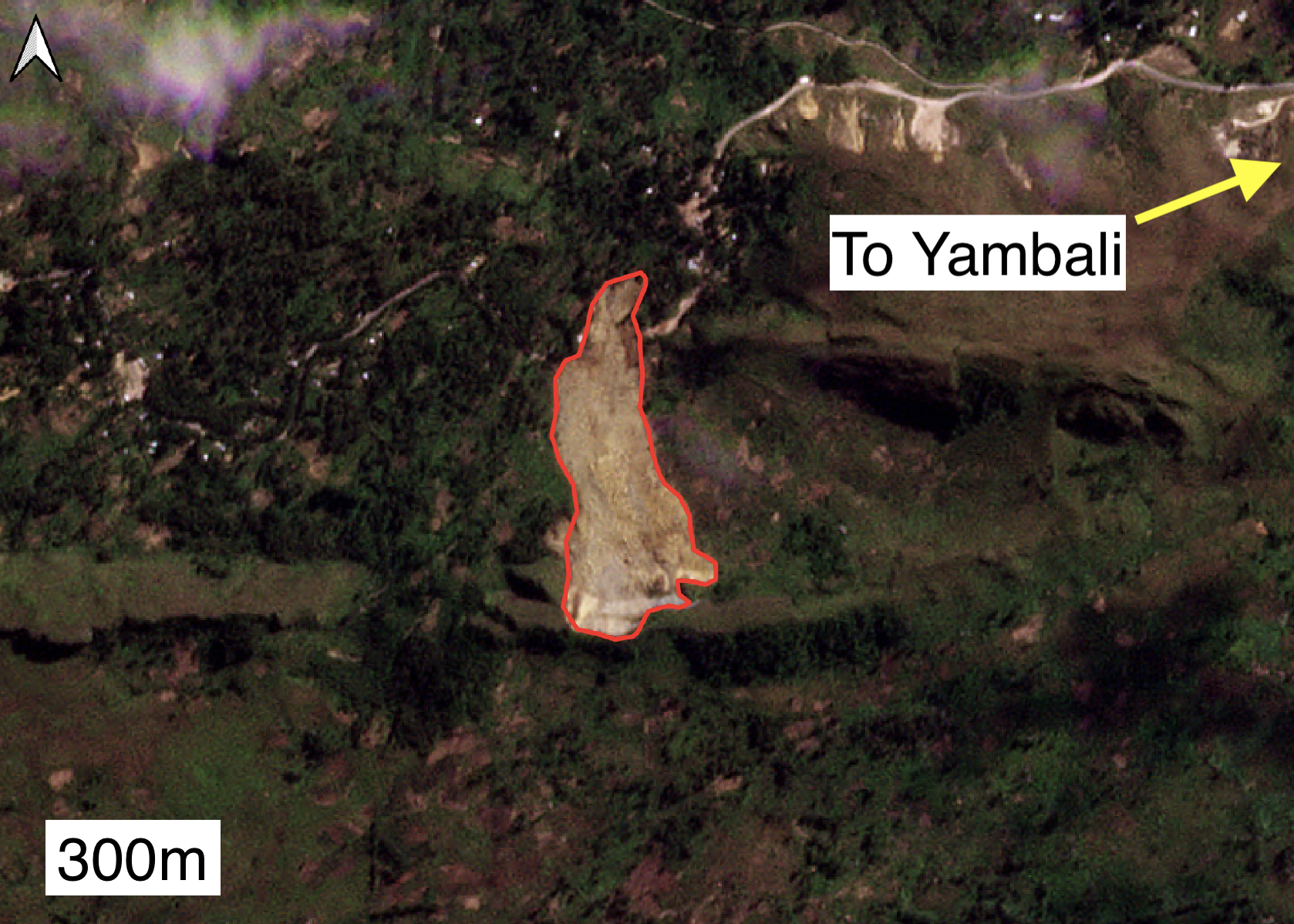}
    \caption{May 26, 2024}
    \end{subfigure}
    
    \caption{PlanetScope (optical) imagery \cite{planet_data} of before and after the landslide in Papua New Guinea. The red polygon shows the extent of the landslide (mapped manually by the authors).}
    \label{fig:landslide_opt}
\end{figure}

\begin{figure}[!t]
    \centering
    
    \begin{subfigure}{.49\linewidth}
    \centering
    \includegraphics[width=\linewidth]{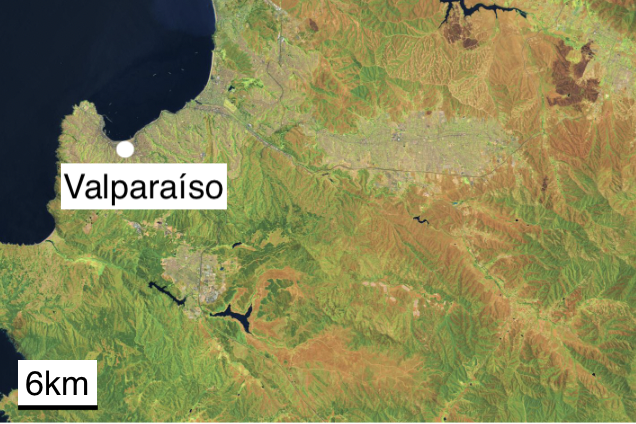}
    \caption{December 20, 2023}
    \end{subfigure}
    \hfill
    \begin{subfigure}{.49\linewidth}
    \centering
    \includegraphics[width=\linewidth]{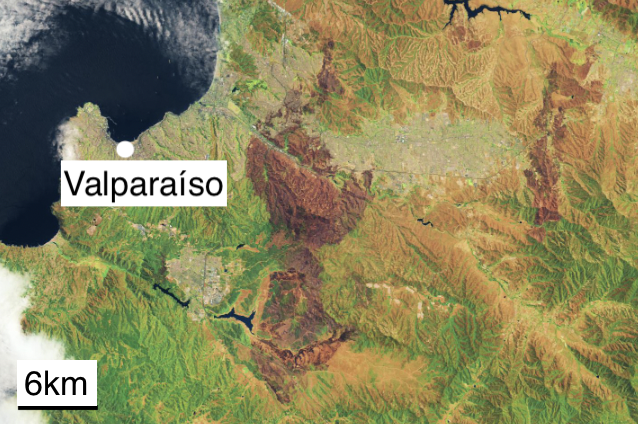}
    \caption{February 5, 2024}
    \end{subfigure}
    
    \caption{Landsat 8 and 9 false color imagery (bands 6, 5, 3) of before and after the fires in the Valpara\'iso region of Chile. The above imagery is taken from NASA Earth Observatory \cite{nasa_inferno_scars_valparaiso}.}
    \label{fig:fire_opt}
    
\end{figure}

\subsection{Mathematical Formulation of the Model}
\label{sec:math_model}

Let $X := \{x_1, x_2, ..., x_T, x_{T+1}\}$ be a sequence of coregistered SAR images, with each $x_i \in \mathbb{R}^{C \times H \times W}$, where $C$ is the number of channels and $H$ and $W$ are the height and width in pixels, respectively.
We call the images $x_1, ..., x_T$ the set of baseline imagery, or pre-images.
The image $x_{T+1}$ represents the latest acquisition image, or post-image.
The input to the model is the set of baseline images, and the output is an estimate of the per-pixel distribution of $x_{T+1}$.
For our work, the channel dimension $C$ will be $2$ as our input imagery is dual-polarization, i.e. VV and VH (co-polarized data VV is vertical transmit and vertical receive and cross-polarized data VH is vertical transmit and horizontal receive).
The spatial dimensions are $H = W = 16$. 

Let $f$ be the model 
parameterized by weights $\theta$. 
Our model seeks to learn the parameters of a normal distribution at each pixel at timestep $T+1$ from the set of baseline imagery. More precisely:

\begin{equation}
\mu_{T+1}, \sigma_{T+1} = f_{\theta}(x_1, x_2, ..., x_T),
\end{equation}
where $\mu_{T+1}, \sigma_{T+1} \in \mathbb{R}^{C \times H \times W}$. 
Since our forecast is probabilistic in nature, we minimize the 
negative log-likelihood loss to find the optimal parameters $\theta^*$ on samples from our training set $D$, consisting of sequences of variable-length baseline imagery:
\begin{equation}\label{loss}
\begin{aligned}
\theta^* & = \\
& \argmin_{\theta} \sum_{X \in D} \Bigl( \frac{1}{2} \text{log}(2\pi \sigma_{T+1}^2) + \frac{(x_{T+1} - \mu_{T+1})^2}{2\sigma_{T+1}^2} \Bigr).
\end{aligned}
\end{equation}
This loss assumes (a) that each polarization is independent and (b) that each pixel in the image $x_{T+1}$ is independent.
Assumption (a) is made because the anti-diagonal components of the covariance matrix of the dual-polarization data typically are small and thus numerically unstable, in part because the two polarizations are correlated over some land cover types and anti-correlated in others.
Assumption (b) is made so we can write down the negative log-likelihood for each training sequence coming from $D$ at once \cite{stephenson2021deep}.
We note that, while we assume each pixel in $T+1$ is independent, all the pixels in the set of baseline images $x_1, x_2, \dots, x_T$ are used in any given estimate of a pixel's distribution in $x_{T+1}$.
Equation \ref{loss} can be optimized using any standard optimization routine; we use the Adam optimizer \cite{kingma2014adam}. 
Note that the loss is computed with respect to the $T+1$ image, and requires no label information; the training process is entirely self-supervised.

\subsection{The Disturbance Metric and Disturbance Mapping}
\label{sec:metric}

\begin{figure*}[!t]
    \centering
    \begin{subfigure}[t]{.705\linewidth}
        \centering
        \includegraphics[width=\linewidth]{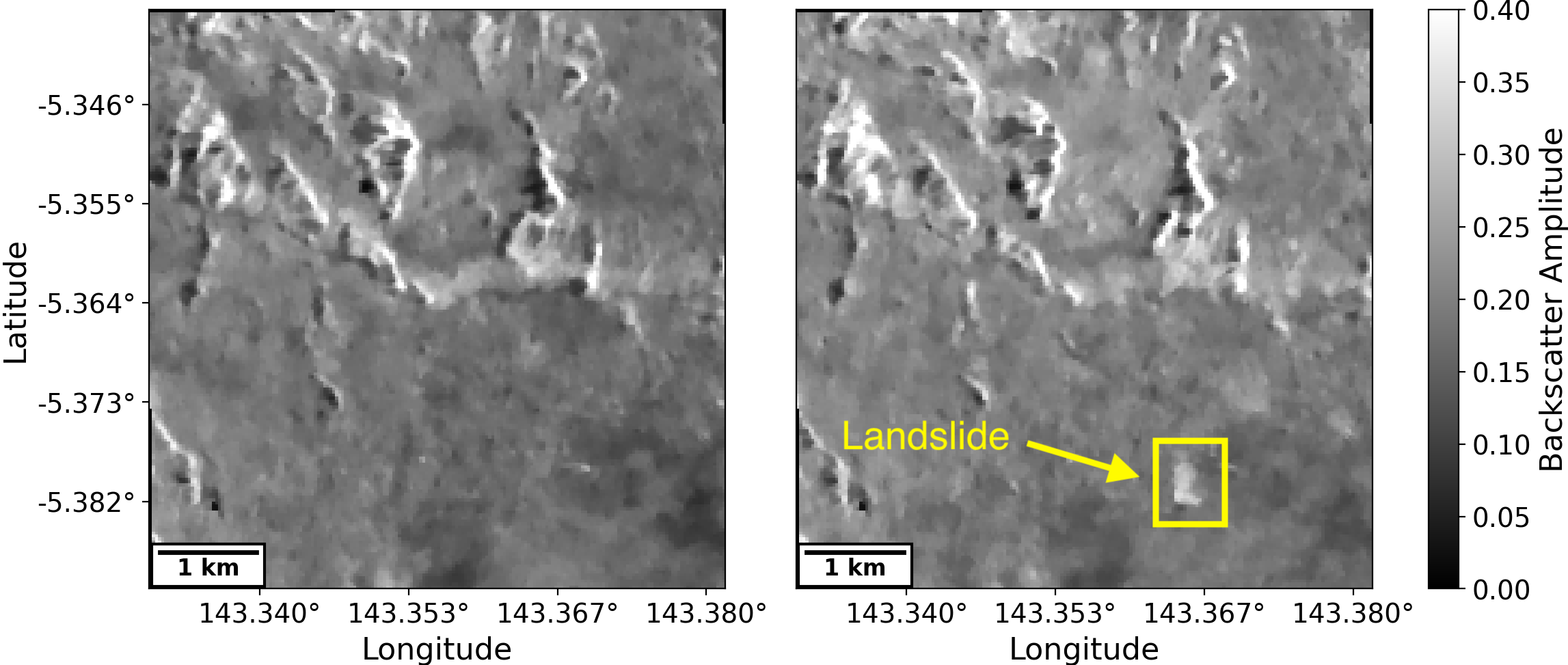}
    \end{subfigure}
    \hfill
    \begin{subfigure}[t]{.28\linewidth}
        \centering        \includegraphics[width=\linewidth]{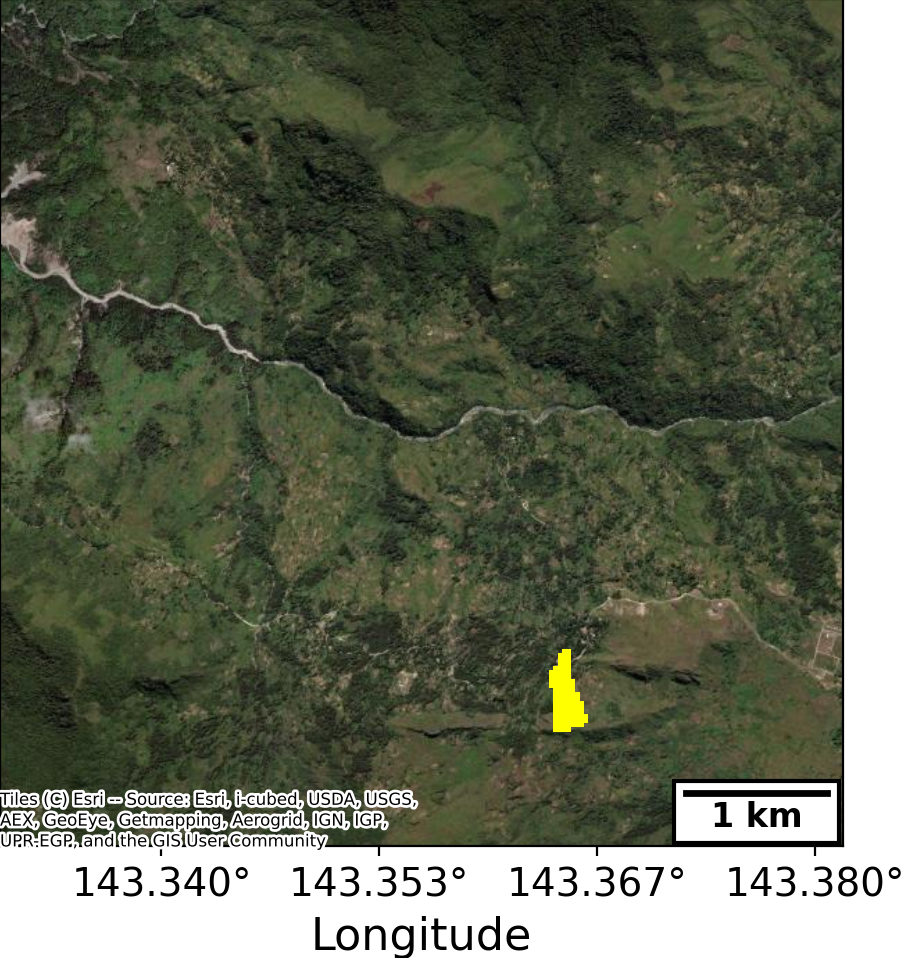}
    \end{subfigure}
    \caption{\textbf{Left}: Pre-landslide SAR imagery (VV), May 22 2024. \textbf{Middle}: Post-landslide SAR imagery (VV), June 3 2024. The landslide is seen in the bottom right corner. \textbf{Right}: Target (ground truth) damage map overlaid on ESRI Composite World Imagery \cite{esri_world_imagery}. The damaged area is shown in yellow. Note that the ESRI imagery is a composite and thus does not capture the recent landslide, and is used as a reference only.}
    \label{fig:landslide_sar}
\end{figure*}

In the previous subsection, we detailed how to take a sequence of baseline images $x_1, ..., x_T$ to estimate the per-pixel distribution.
We can now define a disturbance metric, $d_p$, that provides a per-pixel measurement of disturbance by comparing the estimated distribution to the most recent acquisition $x_{T+1}$. Consider a single polarization $p$, $p =$ VV or VH.
We define our disturbance metric as the 1-dimensional Mahalanobis distance:
\[
d_p = \frac{|x^p_{T+1} - \mu_p | }{\sigma_p}.
\]
This metric can be interpreted as the number of standard deviations from the estimated mean $x_T$.
The larger $d_p$ is, the more likely disturbance has occurred, where disturbance is with respect to the baseline imagery.
A similar metric was used in \cite{stephenson2021deep}, where the absolute value is omitted and referred to as the z-score; however, that work studied interferometric coherence, where a decrease is most important for damage assessment, and so sign must be preserved.
In our case, we consider absolute deviation, since increases or decreases in backscatter could indicate disturbance.

We then combine each polarizations taking the maximum of $d_p$, i.e. $d = \max_p d_p$.
Taking the maximum over $d_p$ effectively means that our delineations consider disturbances in \emph{either} polarization, VV or VH.
One could also consider combining the metrics via a minimum (analogous to a logical ``and'' over the channels) or via addition (a 2-dimensional Mahalanobis distance assuming channel independence).
We elected the maximum (analogous to a logical ``or") as disturbances are often more detectable in one channel than the other.
Indeed, many disturbance related works focus on a single polarization \cite{rs14061449_nava_unet_landslide, rignot1993change}.

A summary of our approach is shown in Fig. \ref{fig:metric-flow}.
To map disturbance, we threshold $d > \tau$ for some positive real number $\tau$.
For the rest of this paper, we call $d$ the \emph{transformer metric}.
Our metric has a clear probabilistic interpretation that a metric exceeding some $\tau$ has an associated probability.
For example, if we assume normality of a pixel, we know that $d_p > 3$ occurs less than 1\% of the time.
In our disaster data, we found thresholds $\tau$ in the range of $3$-$5$ visually suitable for disturbances delineations.
We will systematically analyze thresholds $\tau$ using optically-derived validation data in Section \ref{sec:results}.

Throughout our experiments, 
we will compare to an RNN-based model with the same disturbance metric, and a log ratio-based metric using $\ell$ in Section \ref{sec:disturbance_mapping}, which takes the absolute deviation in dB with respect to a particular polarization.
We again combine polarizations via the maximum and set a baseline $I_0$ by taking a per-pixel median of the pre-event imagery.

\subsection{Model Architecture, Hyperparameters, and Training}\label{sec:model_params}

\begin{figure*}[!t]
    \centering
    \begin{subfigure}[t]{.71\linewidth}
        \centering
        \includegraphics[width=\linewidth]{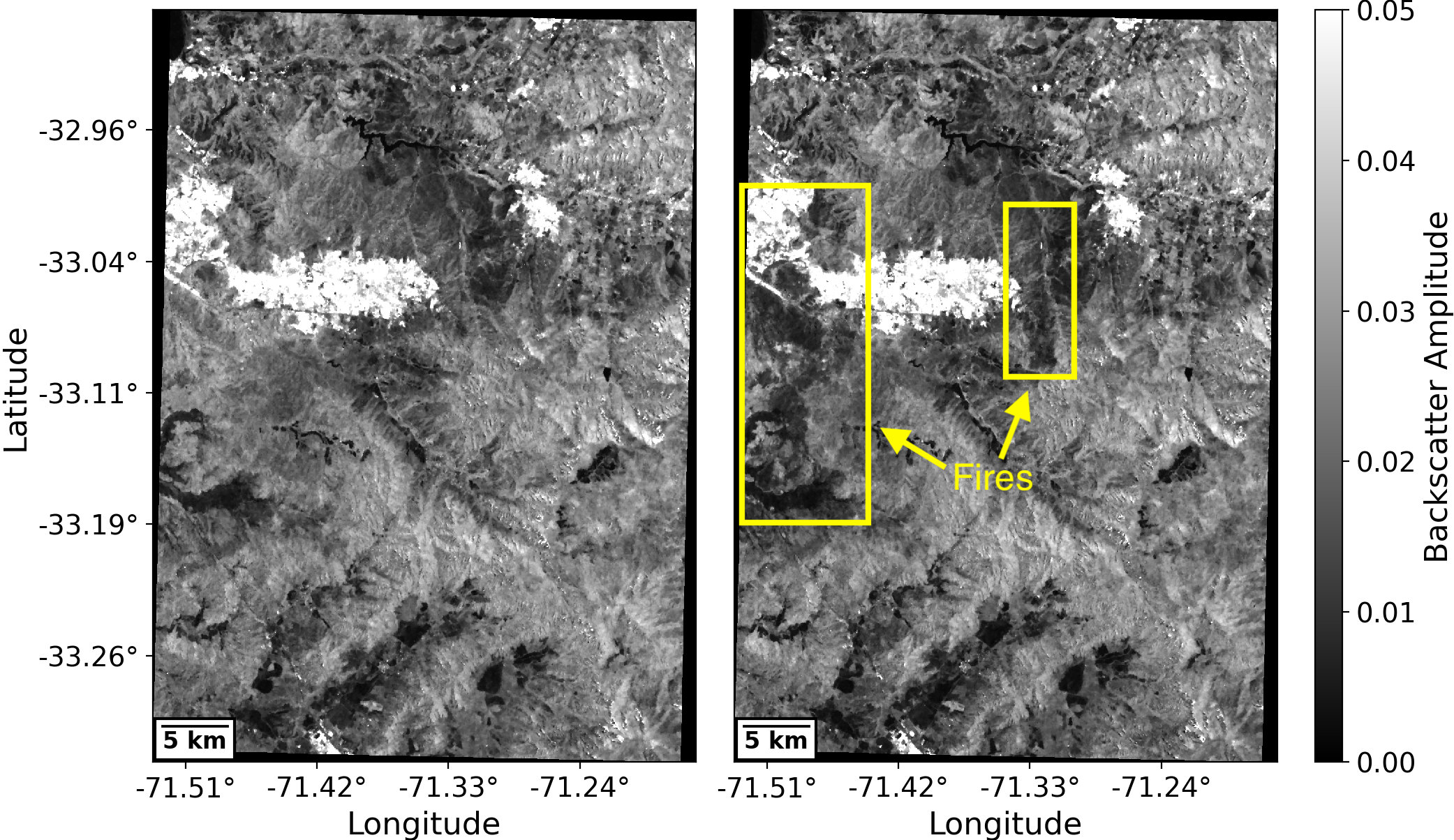}
    \end{subfigure}
    \hfill
    \begin{subfigure}[t]{.275\linewidth}
        \centering
        \includegraphics[width=\linewidth]{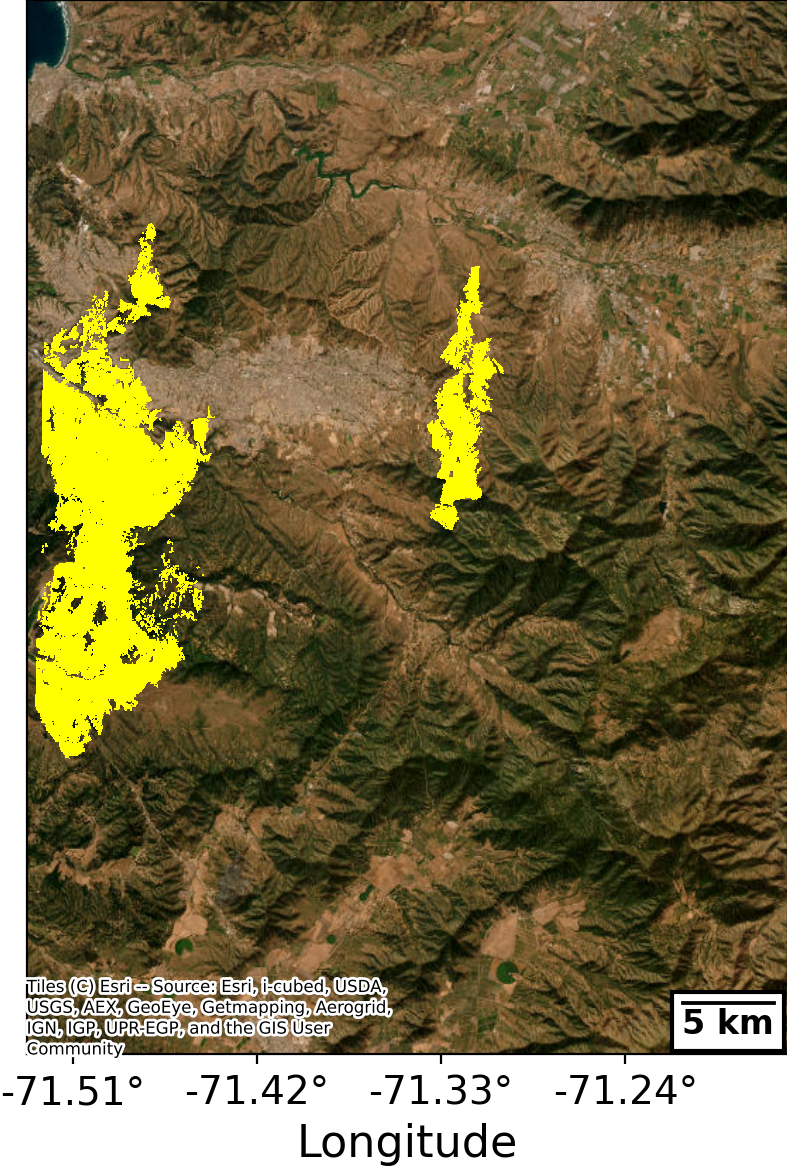}
    \end{subfigure}
    \hfill
    \caption{\textbf{Left}: Pre-fire SAR imagery (VH), January 23 2024. \textbf{Middle}: Post-fire SAR imagery (VH), February 28 2024. \textbf{Right}: Target (ground truth) damage map from Copernicus overlaid on ESRI World Imagery \cite{esri_world_imagery}. The damaged areas are shown in yellow. Note that the ESRI world imagery  is a composite and thus does not capture the recent fires, and is used as a reference only.}
    \label{fig:fire_sar}
\end{figure*}

We use an encoder-based vision transformer architecture. We take patches of the input images and add learned spatial and temporal positional embeddings, as done previously for spatiotemporal data in \cite{tong2022videomae, feichtenhofer2022masked, cong2022satmae}.
We elected not to use more sophisticated spatial-temporal encodings (see e.g. \cite{tseng2023lightweight, reed2023scale, clayfoundation2024model, jakubik2023foundation}) as we wanted our model to be simple to illustrate the efficacy of this type of approach to disturbance mapping.
We will explore more sophisticated spatiotemporal positional encoding in the future.

Our transformer encoder uses a model dimension of 256, 4 attention heads, 4 layers, and a feedforward dimension of 768. 
The input to the model is a sequence of $T$ SAR images, each with size $2 \times 16 \times 16$ (channels, heigh, and width, respectively, for a total input size of $T \times 2 \times 16 \times 16$. The channel dimension $2$ is due to the dual-polarization SAR imagery (VV/VH). In each batch, $T$ is chosen randomly to be between 2 and 10 so the model can learn to forecast the distributions with varying amounts of pre-event information. 
Operationally, this is important as data may often be missing or an area of interest lacks a long acquisition history.
For example, Sentinel-1 data does not consistently image in 12-day increments, so in some cases many months can go by before additional data is collected. 
Our validation datasets discussed in Section \ref{sec:results} have 10, 7, and 4 pre-event images available, underscoring the need for a temporally flexible model. Future work will focus on scaling up the model to allow for larger values of $T$.

Each $2 \times 16 \times 16$ SAR image is ``patchified'' into $2 \times 8 \times 8$ patches with a learned spatial and temporal embedding. 
For example, for a sequence of ten $16 \times 16$ SAR images, the input sequence to the transformer will have 40 ($4$ patches $\times$ $10$ temporal steps) elements. 

We use a dropout of 0.2 and ReLU activations. 
After the transformer learns representations of the input sequence, we use a 2 layer, fully connected (with hidden dimension 768 and a ReLU activation) projection head to output the predicted mean and standard deviation of each pixel in each channel in the subsequent image (both the mean and the standard deviation predictions have their own projection head). This results in a model with approximately 3.3 million parameters. 

Previous work on deep learning-based disturbance mapping on SAR used an RNN Gated Recurrent Unit (GRU) \cite{cho2014properties, stephenson2021deep}. 
For a fair comparison, we construct a GRU with a nearly identical parameter count as the transformer (we note that the original work \cite{stephenson2021deep} applied a GRU to coherence values from SAR, whereas our work uses backscatter).
To construct the GRU architecture with 3.3 million parameters, we use a model dimension of 326, 4 layers, a hidden dimension of 978 for the projection head, and the same dropout value, 0.2. This model also uses projection heads for the mean and standard deviation predictions. Since transformers utilize patches with positional embeddings, it is not obvious whether a fair comparison to an RNN should have an input size of $8 \times 8$ or $16 \times 16$. We found that the input size of $8 \times 8$ performed better (although still surpassed by the transformer), which is the model we report results for throughout Section \ref{sec:results}. For completeness, we include results for the $16 \times 16$ input size in Section \ref{sec:ablation}.

The training of both models use a batch size of 256, and train for 50 epochs using Adam \cite{kingma2014adam} with an initial learning rate of $0.0001$, which then decays to $0.00001$ after 25 epochs. All experiments were run on a single NVIDIA A100 GPU with an 80GB memory.



\subsection{Model Inference}\label{sec:inference}

For model inference, we often have much larger images than $16 \times 16$; a single burst RTC-S1 product is approximately $1000 \times 3000$ pixels.
To apply our model, we sweep it across the image (for computational reasons, we use a stride of 4 for the fire and flood delineations, see Section \ref{sec:results}) and average the estimates generated at each pixel.
Without this sweeping approach, the predictions leave ``edge artifacts" because the borders of each $16 \times 16$ window have less neighboring pixels to inform the estimates.
Indeed, inference of our small windows exhibit the so-called receptive field phenomena \cite{luo2016understanding_receptive_field}.

\section{RTC-S1 Data}
\label{sec:data}

The OPERA RTC-S1 provides measurements of the radar backscatter, $\gamma^0$, corrected for radiometry and topography \cite{small_rtc, shiroma2023opera}. 
All RTC-S1 products can be accessed through the Alaska Satellite Facility Distributed Active Archive Center (ASF DAAC) \cite{opera_rtc_s1_v1_daac}.
These corrections result in an analysis-ready product that primarily captures signals related to the physical characteristics of ground-scattering surfaces, such as surface roughness, soil moisture, and vegetation structure. 
The RTC-S1 is generated from Copernicus Sentinel-1 (S1) interferometric wide (IW) single-look complex (SLC) data, provided by the European Space Agency (ESA). 
Sentinel-1 constellation can currently acquire data at 12 day repeat pass temporal frequency, i.e. with respect to a fixed orbit track.
The ground sampling when independent of orbit track is increased to between 1 and 6 days; this combines ascending (satellite flying north and looking east) and descending (satellite flying south and looking west) geometries. 
The imaging mode and temporal coverage of Sentinel-1 data varies with location on Earth \cite{asf_sentinel1_acquisition_maps}. 
The RTC-S1 product has a near-global geographic scope that covers all landmasses, excluding Antarctica, and its temporal frequency aligns with the availability of Sentinel-1A SLC data. 
RTC-S1 currently only includes data from Sentinel-1A satellite because OPERA started forward mode production in October 2023 after the loss of Sentinel-1B. 
Each RTC-S1 product corresponds to a single Sentinel-1 burst and is mapped onto standardized Universal Transverse Mercator (UTM) or Polar Stereographic projection systems. 

RTC-S1 uses the Copernicus global 30 m (GLO-30) Digital Elevation Model (DEM) as the reference for topographic correction and geocoding. 
The data are in GeoTIFF format, and have a spatial resolution of 30 meters. 
The RTC-S1 data also includes layover and shadow masks as well as bistatic and tropospheric corrections to improve data quality. 
Static layer files such as incidence angle, local incidence angle, masks, and number of looks are used. 
The OPERA RTC-S1 and static layers are generated using the InSAR Scientific Computing Environment version 3 (ISCE3) software developed at JPL/Caltech \cite{rosen2018insar}. 
Additionally, the RTC-S1 serves as the foundational dataset for other OPERA products including the Dynamic Surface Water eXtent from Sentinel-1 product (in production since Sep. 2024) and the future OPERA Disturbance from Sentinel-1 product (expected production in 2026). 
All of the OPERA products are designed to address the needs of U.S. Federal Agency stakeholders identified by the Satellite Needs Working Group.

\begin{figure*}[!t]
    \centering
        \begin{subfigure}[t]{.7075\linewidth}
        \centering
        \includegraphics[width=\linewidth]{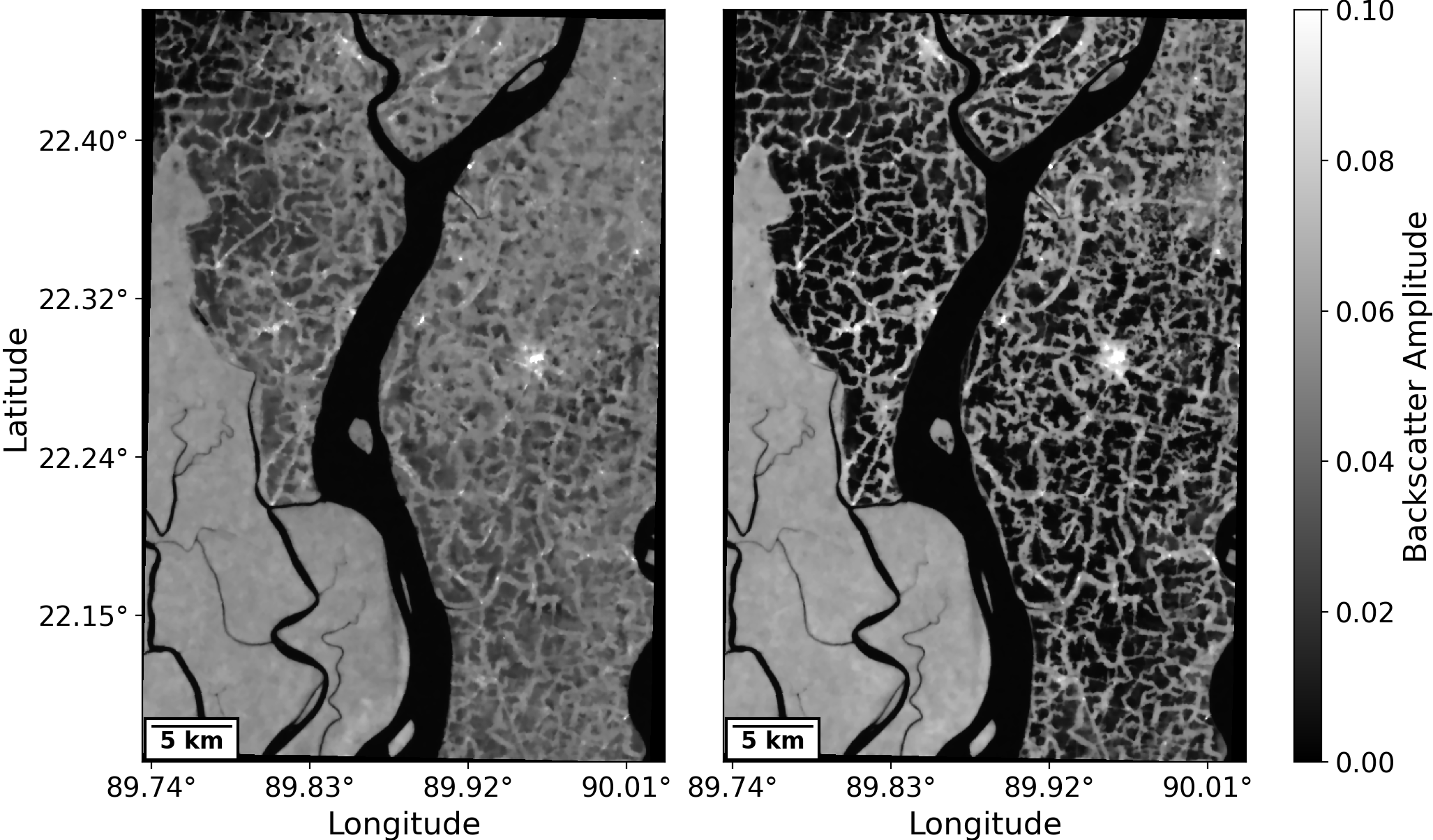}
    \end{subfigure}
    \hfill
    \begin{subfigure}[t]{.28\linewidth}
        \centering
        \includegraphics[width=\linewidth]{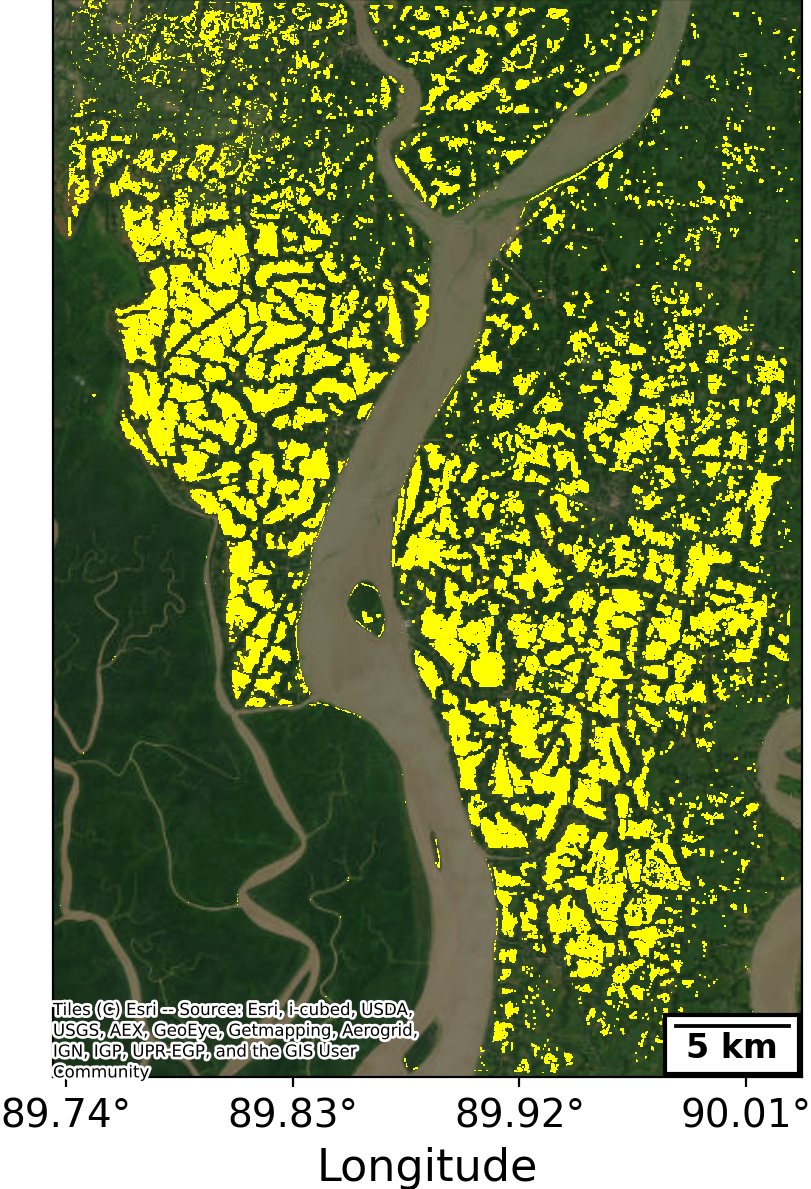}
    \end{subfigure}
    \caption{\textbf{Left}: Pre-flooding SAR imagery (VH), May 18 2024. \textbf{Middle}: Post-flooding SAR imagery (VH), July 5 2024. The flooding is seen via the reduction of the backscatter all around the Baleshwari River. \textbf{Right}: Target (ground truth) damage map from UNOSAT, overlaid on ESRI World Imagery \cite{esri_world_imagery}. The damaged areas are shown in yellow. Note that the ESRI imagery is a composite and thus does not capture the recent flooding, and is used as a reference only. We note that flooding continued throughout the summer, and we use data collected on July 5 as our post-event imagery to align with the damage map from UNOSAT.}
    \label{fig:flood_sar}
\end{figure*}

\subsection{Training Data}

The training data is comprised of $2,511,348$ sequences of 11 SAR images from OPERA RTC-S1 (i.e., $T = 10$ is the largest possible value for $T$ in training), which are each 2 channel (VV and VH) and $16 \times 16$ pixels. Future work will focus on increasing the size of the dataset and the temporal length of the sequences $T$. This training data is from OPERA RTC-S1 data that is currently not in the ASF archive, but rather stored in a public s3 bucket for the OPERA project validation \cite{opera_rtc_s1_table_for_validation}.
The data is used to overlap the sites and validation period (October 2020 - October 2022) used by OPERA DIST-HLS validation \cite{opera_validation_plan, opera2024distvalidation}.
The data collection includes imagery from Sentinel-1B so some of the temporal sampling occurs with frequency of 6 days.

\subsection{Data Preprocessing}
Before using our model to estimate the distributional parameters, we perform basic SAR despeckling.
We utilize homomorphic despeckling as described in \cite{deledalle2017mulog} using a total variational denoising model \cite{rudin1992nonlinear}. Homomorphic despeckling is total variation denoising on dB-transformed backscatter imagery that, after despeckling, is inverted back into $\gamma^0$.

The probabilistic framework of our approach approximates the unknown pixel distribution using a Gaussian, which has domain $\mathbb{R}$. 
SAR $\gamma^0$ backscatter is within $[0,1]$ so we follow the same pre-processing step as in \cite{stephenson2021deep}, taking the logit transform of the backscatter data.
The logit is a bijection of $[0,1]$ to $\mathbb{R}$.
This is done on the despeckled imagery.
\begin{equation}\label{eq:logit}
\text{logit}(x) = \text{log}\Bigl( \frac{x}{1-x} \Bigr)
\end{equation}
For clarity, all figures in this paper are presented in their original backscatter range $[0,1]$.
These two preprocessing steps are performed before application of the model described above.

\section{Disaster Data}
\label{sec:disaster_data}

In this section we discuss the data used to evaluate our disturbance delineations.
All the data was derived either manually by the authors (in the case of the Papua New Guinea Landslide) or by external disaster response projects such as UNOSAT or Copernicus Emergency Management Services (CEMS).
We have curated these delineations and more events at this GitHub repository \cite{opera_dist_s1_events}, along with provenance.
The Bangladesh flood was derived from Sentinel-1 data using an image pair by UNOSAT \cite{hdx_water_extents_bangladesh_2024}.
The other disaster delineations (landslide and fire) are derived from VHR optical data (i.e. with spatial resolution $\leq$ 3m).
Evaluating disturbance mapping with these ground truth datasets presents some limitations.
First, we expect some inherent differences between disturbance maps derived from optical and SAR data. However, optical and SAR data tend to exhibit high agreement in these cases when dense vegetation is cleared by a fire or a landslide \cite{woodhouse2017introduction}.  Second, real disturbances unrelated to the natural disasters are not accounted for in this setting, but the abrupt nature of disasters means we expect this to have a negligible impact on our results. Overall, these natural disaster datasets serve as a high-quality ground truth for evaluating our methodology.

\begingroup
\makeatletter
\long\def\@makecaption#1#2{%
\begin{center}
  {\normalfont#1}\\{\MakeUppercase{\normalfont #2}}
\end{center}
}
\makeatother

\begin{table*}[!t]
\caption{Summary of Results}
\centering
\begin{tabular}{||c||c|c|c|c|c|c||}
\hline
 & \multicolumn{2}{c|}{\textbf{Landslide}} & \multicolumn{2}{c|}{\textbf{Fire}} & \multicolumn{2}{c||}{\textbf{Flood}} \\ 
\hline
 \textbf{Model} & PR AUC & Max $F_1$ Score & PR AUC & Max $F_1$ Score & PR AUC & Max $F_1$ Score \\ 
\hline
Transformer & \textbf{0.732} & \textbf{0.769} & \textbf{0.680} & \textbf{0.645} & \textbf{0.754} & \textbf{0.701} \\
\hline
RNN         & 0.642 & 0.699 & 0.624 & 0.596 & 0.705 & 0.661 \\
\hline
Log Ratio   & 0.067 & 0.194 & 0.391 & 0.474 & 0.715 & 0.655 \\
\hline
\end{tabular}
\label{tab:results}
\end{table*}

\endgroup

\subsection{2024 Papua New Guinea Landslide}\label{sec:landslide_data}

On May 24, 2024, a large catastrophic landslide occurred near Yambali, Enga Province, Papua New Guinea (see Fig. \ref{fig:landslide_opt}) \cite{landslide_cnn, abc_news_png_landslide}. 
News media reported that that the landslide buried numerous villages, with the death toll estimated to be between 650 and 2,000 people \cite{landslide_cnn}; the official death toll at time of writing is 670 \cite{landslide_un}. In addition, approximately 7,000 people have been displaced by the event \cite{landslide_cnn}.

Our input to the model is a sequence of ten pre-event OPERA RTC-S1 SAR images from 1/23/24 to 5/22/24 at 12 day intervals (except 5/10/24 when there was no data distributed despite a Sentinel-1 pass) and a post-event image from 6/3/2024. 
To delineate disturbances, we compute the Mahalanobis distance between the models' estimate distribution and this post-event image as described in Section \ref{sec:metric}.
We analyze an area of approximately 6 km $\times$ 6 km. We manually mapped the landslide extent using PlanetScope optical imagery (3 m pixel) \cite{planet_data}, see Fig. \ref{fig:landslide_sar}. 




\subsection{2024 Valpara\'iso, Chile Wildfire}

In early February of 2024, a series of wildfires began in the Valpara\'iso region of Chile \cite{chile_nasa}. 
Exacerbated by the historic drought in the region and the 2023 El Ni\~no weather event, by February 3 the fires had burned over 8,500 hectares \cite{chile_nasa}. The death toll is at least 131 people \cite{fire_ap}. 
Fig. \ref{fig:fire_opt} shows Landsat 8 and 9 false color imagery of two burn scars\cite{nasa_inferno_scars_valparaiso}.

For this site, The Copernicus Emergency Management Service (CEMS) \cite{CEMS} produced a disturbance map of the region that we use as our ground truth, see Fig. \ref{fig:fire_sar}. 
The area of interest is approximately 45 km $\times$ 33 km.
The data was generated by a variety of VHR sources and was continuously updated for several months after the fires started \cite{esri_storymaps_valparaiso}.
Our RTC-S1 data used included all seven available pre-event SAR images from 11/12/24 to 1/23/24 (all Sentinel-1 passes resulted in distributed data) and one post-event SAR image from 2/29/24. 
Although there was data collected and distributed on February 5 and 17 by Sentinel-1 as the fires were ongoing, there was noticeable increase in disturbance in the February 29 data. This acquisition was also closer in time to the VHR map, which was updated well beyond the fire's start \cite{esri_storymaps_valparaiso}.

\subsection{2024 Bangladesh Floods}

On May 26, 2024, Cyclone Remal landed in Bangladesh, causing major flooding that has continued throughout the summer \cite{bangladesh_unicef}. 
At least 4.6 million people were affected throughout the summer \cite{bangladesh_unicef}. 
Because the flooding occurred throughout the summer, it was challenging to establish a baseline of imagery in the RTC-S1 archive in which these floods were not present.
We selected validation data containing 4 pre-event images from entirely before the cyclone, acquired between 3/31/24 and 5/18/24.
We use a ground truth damage map from the United Nations Satellite Center (UNOSAT) \cite{UNOSAT} which was derived from a Sentinel-1 image on July 4th \cite{hdx_water_extents_bangladesh_2024}.
We use the nearest RTC-S1 post-event image, which was acquired July 5th, 2024.
Our flooding analysis is a subset of the country along Ganges delta around the Baleshwari River. The area of interest is approximately 42 km $\times$ 30 km, see Fig. \ref{fig:flood_sar}.

\section{Results}\label{sec:results}

\subsection{Evaluation Procedure and Performance Metrics}

We evaluate and compare the performance of our proposed self-supervised transformer, the RNN from \cite{stephenson2021deep} adapted for use with RTC-S1, and the classical log ratio approach with per-pixel medians from the baseline set \cite{rignot1993change}.
The definition of these metrics and how they utilize dual-polarization imagery is found in Section \ref{sec:metric}.

For our evaluation procedure, our baseline imagery is established from the dates in Section \ref{sec:disaster_data}. However, we set aside the last pre-event image (i.e. the last available image before the event).
The post-event image and this final pre-event image are then used to assess the disturbance delineations.
We assume 1) the final pre-event image has \emph{no} disturbance and 2) the post-event image matches the external disturbance delineation. This way, we evaluate each method on its ability to accurately delineate disturbances while also not raising false positives when we expect no disturbances. This is an important consideration in practice. In a disaster response scenario, false positives can be extremely costly; resources could be allocated to areas where they are not needed at the expense of areas that do.

\begin{figure*}
    \centering
    \begin{subfigure}{\linewidth}
        \centering
        \caption{Transformer}
        \includegraphics[width=\linewidth]{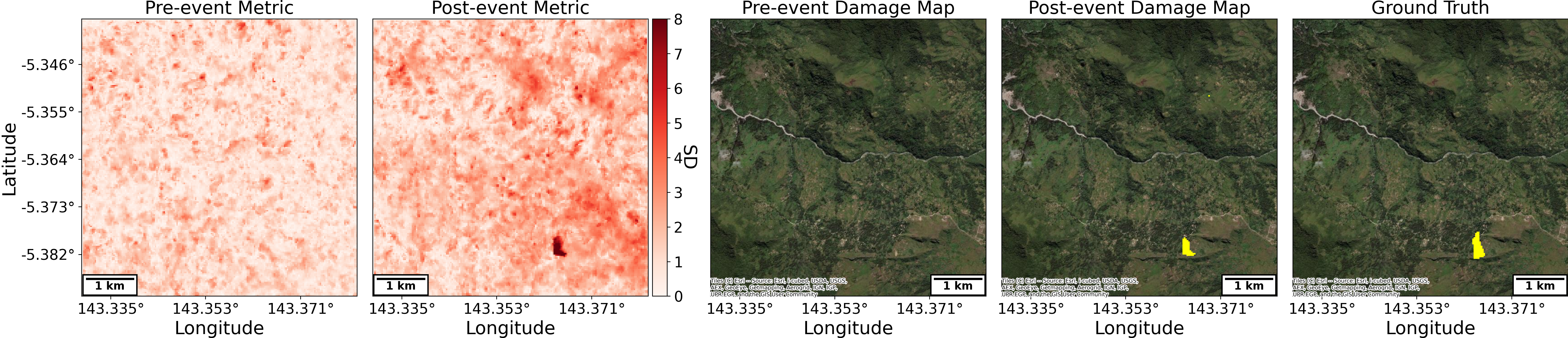}
    \end{subfigure}
    \hfill
    \begin{subfigure}{\linewidth}
        \centering
        \caption{RNN}
        \includegraphics[width=\linewidth]{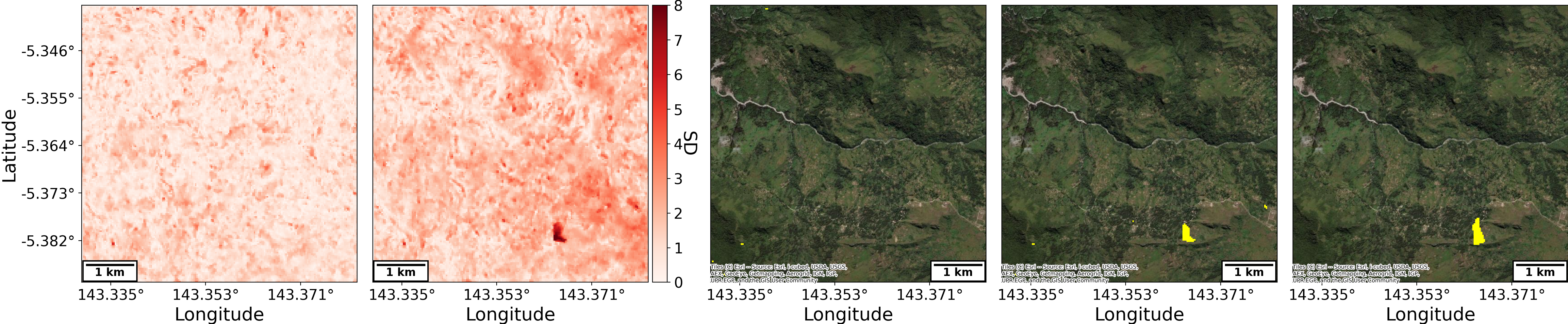}
    \end{subfigure}
    \hfill
    \begin{subfigure}{\linewidth}
        \centering
        \caption{Log Ratio}
        \includegraphics[width=\linewidth]{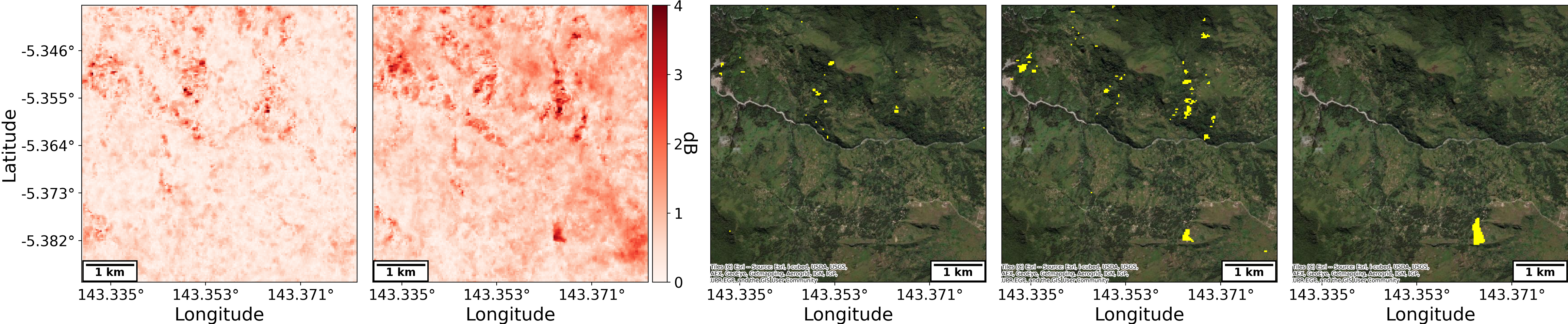}
    \end{subfigure}
    \caption{
    Disturbance metrics, damage maps, and ground truth data from the landslide in Papua New Guinea. Each row corresponds to a different model. The first two columns show the model's disturbance metric (see Section \ref{sec:metric}) before the landslide (first column) and after the landslide (second column). 
    A strong model should produce a metric with low values pre-landslide (when we expect little or no disturbance since no landslide has occurred yet) while also accurately and confidently capturing the damage extents post-landslide. 
    The third and fourth columns show the best (in terms of $F_1$ score, see Fig. \ref{fig:landslide_pr}) binary damage maps derived from each method's metric. 
    The rightmost column shows the ground truth damage map, reproduced from Fig. \ref{fig:landslide_sar} (right panel) for convenience.
    The transformer metric is the strongest, resulting in the damage map with the fewest false positives while also accurately capturing the extents of the landslide.
    }
    \label{fig:landslide_nodamage}
\end{figure*}


Our analysis combines qualitative and quantitative results to facilitate visual comparisons and highlight the strength of our approach. Qualitative results include two map types: (1) disturbance metric maps visualizing $d$ (transformer or RNN) or $\ell$ (log ratio) across the area of interest, and (2) binary (i.e., disturbance or no disturbance) disturbance delineations based on the optimal threshold $\tau$ (according to $F_1$ score, see below). Operationally, ground truth information is not available, so these maps are the critical outputs that delineate disturbances. In our experiments, these maps highlight the transformer's effectiveness in identifying disturbances while limiting false positives.

For our quantitative results, we report Precision-Recall (PR) curves, the associated area under the curve (AUC), and $F_1$ score (harmonic mean of precision and recall) \cite{murphy2012machine}.
Precision is the percentage of pixels predicted as damage by the model that are actually damaged (positive predictive rate), while recall is the percentage of all damaged pixels that are identified (true positive rate). $F_1$ score is then computed by
\begin{equation}
F_1 = 2 \cdot \frac{\text{precision} \cdot \text{recall}}{\text{precision}+\text{recall}}
\end{equation}
The ranges of all these metrics is $[0,1]$, where higher is better. The precision-recall curves are found by varying the threshold $\tau$ (see Section \ref{sec:metric}) and computing the above metrics.
A high quality but imperfect classifier will naturally trade off between precision and recall; 
high thresholds will result in high precision but low recall, and low thresholds will result in low precision and high recall.
Hence, the area under the precision-recall curve (PR AUC) is an overall measurement of classifier quality that accounts for the entire range of $\tau$.
(A perfect classifier will achieve an AUC of 1, while a random classifier will be a horizontal line with a y-intercept equal to the percentage of positive samples in the dataset.) In all of our experiments, our transformer-based approach achieves the highest PR AUC.


 
Accuracy and Receiver Operating Characteristic (ROC) curves are two common metrics which we do not report here, as they are not as meaningful in datasets with significant class imbalances \cite{davis2006relationship}; this is the case in disturbance mapping, where we typically expect large amounts of non-damaged (negative) pixels and smaller amounts of damaged (positive) pixels\footnote{For example, in a scene with 99\% negative and 1\% positive pixels, a naive model which classifies every pixel as negative (and hence not delineate any disturbance extents) would achieve 99\% accuracy, but $0$ recall and undefined precision.}.

\subsection{2024 Papua New Guinea Landslide}

Fig. \ref{fig:landslide_nodamage} shows the transformer metric (see Section \ref{sec:metric}), the RNN metric, and the log ratio metric.
The first column is the final pre-event image and the second column is the post-event image (see Section \ref{sec:landslide_data}). 
The two deep methods, particularly the transformer, are significantly more accurate than the log ratio with respect to our validation data, identifying the landslide extents more confidently and limiting the number of false positives particularly in the first pre-event image. 

\begin{figure}
    \centering
    \includegraphics[width=.85\linewidth]{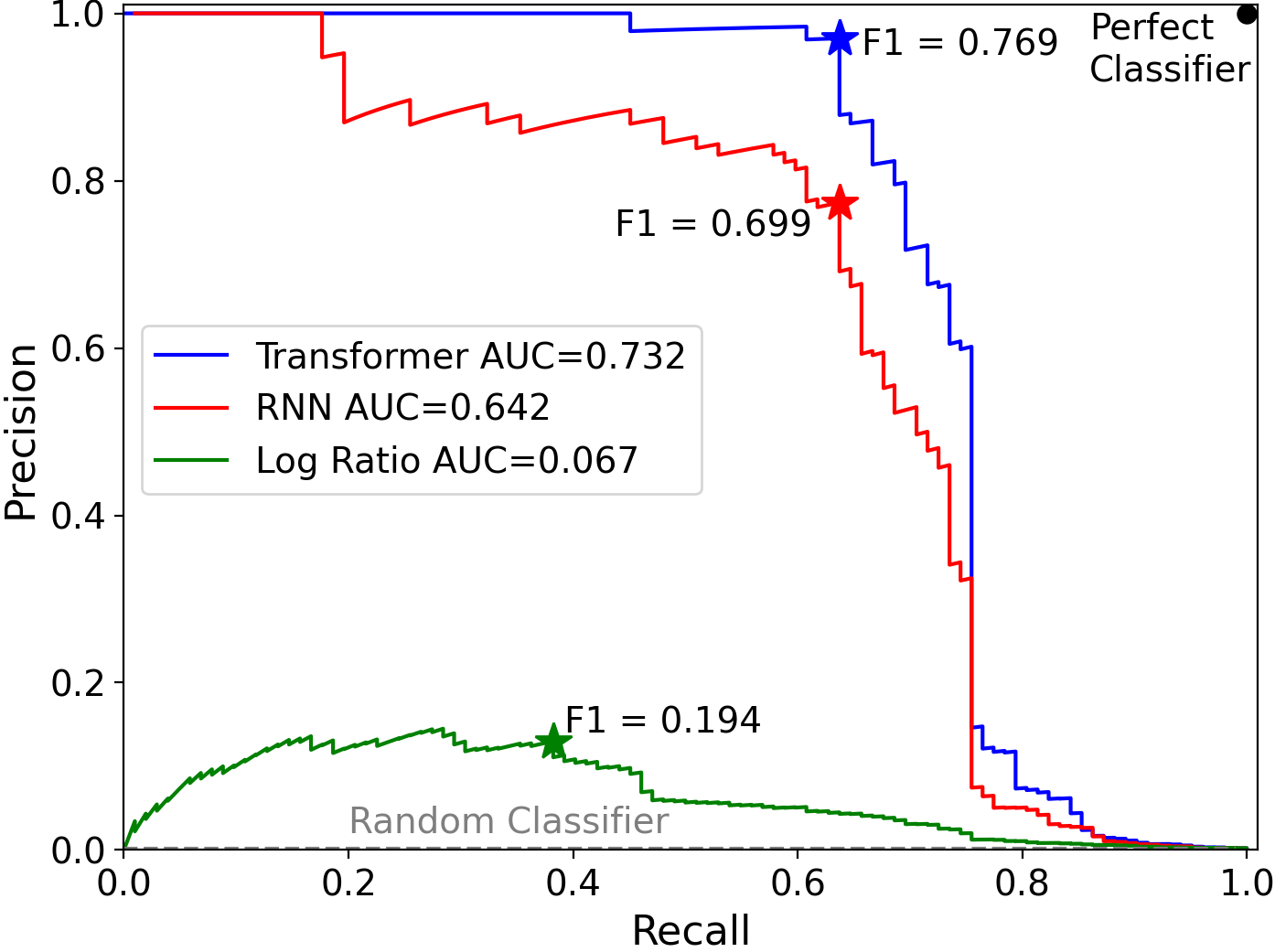}
    \caption{PR curves for the landslide scene. The transformer achieves the largest area under the curve (AUC). The transformer also obtains the best $F_1$ score (starred for each model). 
    The disturbance delineations corresponding to the optimal $\tau$ that obtains optimal $F_1$ score are shown in Fig. \ref{fig:landslide_nodamage}, third and fourth columns.}
    \label{fig:landslide_pr}
\end{figure}

Fig. \ref{fig:landslide_pr} shows the PR curves for the landslide scenes. 
The transformer achieves by far the best AUC ($0.732$) indicating overall model quality independent of threshold $\tau$. 
The RNN is second with an AUC of $0.642$, and both deep methods significantly outperform the classical log ratio method (AUC of $0.067$), which barely outperforms random guessing. 
The optimal $\tau$ and corresponding $F_1$ score is starred.
The transformer obtains the best $F_1$ score of $0.769$, the RNN $0.699$, and the log ratio $0.194$.
The disturbance delineations corresponding to the optimal $\tau$ are shown in Fig. \ref{fig:landslide_nodamage}, third and fourth columns. While in practice, the optimal $\tau$ would not be known \emph{a priori}, we include these figures to highlight an example disturbance delineation from each model.

\begin{figure}
    \centering
    \includegraphics[width=.85\linewidth]{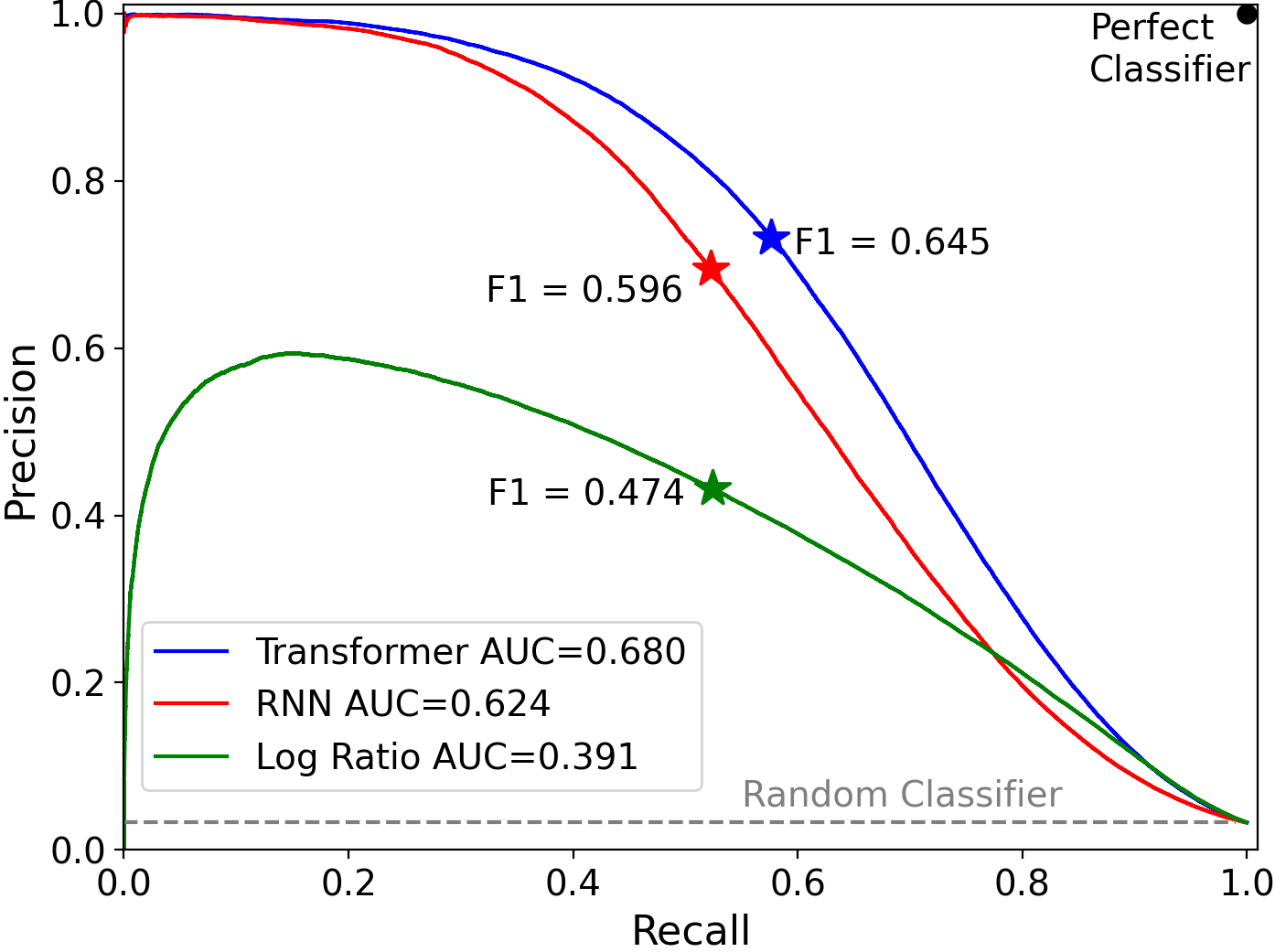}
    \caption{PR curves for the Chilean fire scene. The transformer is the most performant model, achieving the largest area under the curve. The transformer also obtains the best $F_1$ score (starred for each model). The binary damage maps corresponding to these starred $F_1$ scores are shown in Fig. \ref{fig:fire_nodamage}, third and fourth columns.}
    \label{fig:fire_pr}
\end{figure}

\subsection{2024 Valpara\'iso, Chile Wildfire}

\begin{figure*}[!t]
    \centering
    \begin{subfigure}{\linewidth}
        \centering
                \caption{Transformer}
        \includegraphics[width=\linewidth]{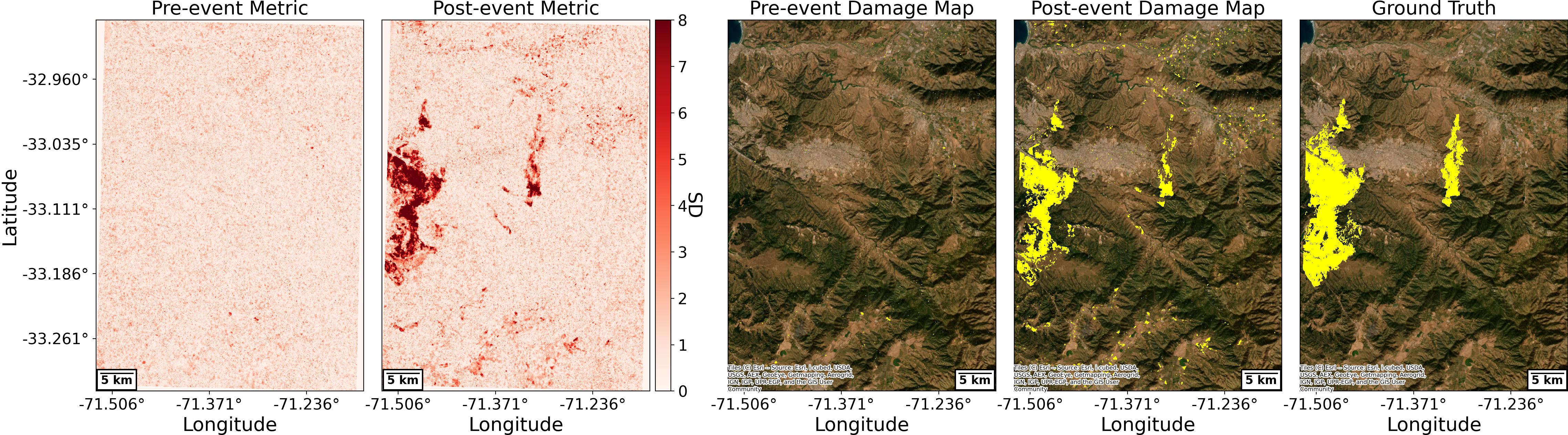}
    \end{subfigure}
    \hfill
    \begin{subfigure}{\linewidth}
        \centering
                \caption{RNN}
        \includegraphics[width=\linewidth]{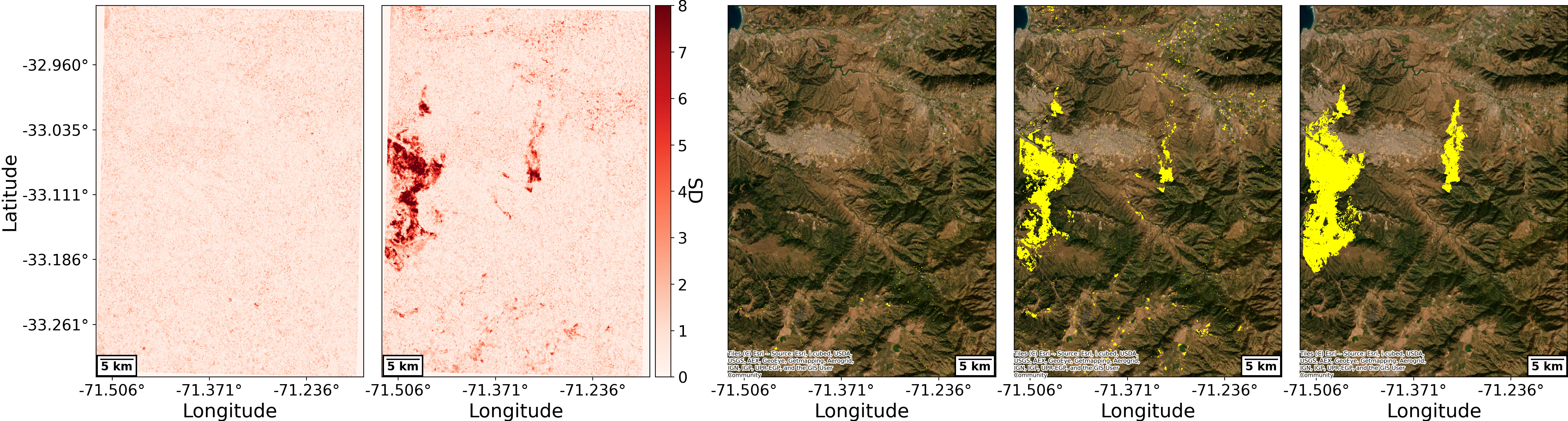}
    \end{subfigure}
    \hfill
    \begin{subfigure}{\linewidth}
        \centering
        \caption{Log Ratio}  \includegraphics[width=\linewidth]{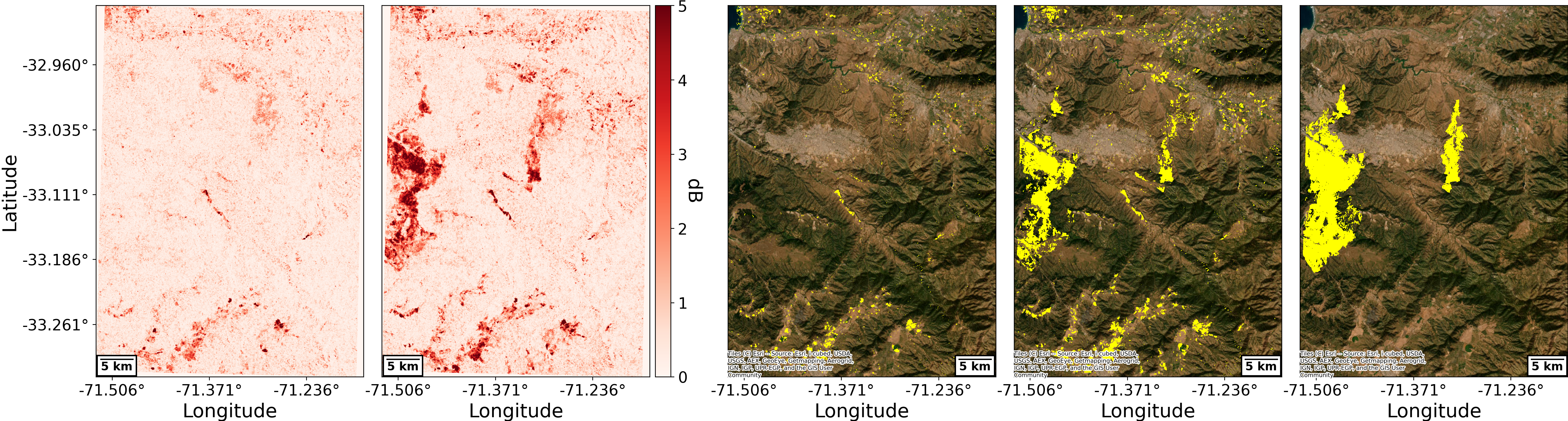}
    \end{subfigure}
    \hfill
    \caption{
    Disturbance metrics, damage maps, and ground truth data from the fire in Chile. Each row corresponds to a different model. The first two columns show the model's disturbance metric (see Section \ref{sec:metric}) before the fire (first column) and after the fire (second column). 
    The third and fourth columns show the best (in terms of $F_1$ score, see Fig. \ref{fig:fire_pr}) binary damage maps derived from each method's metric. 
    The rightmost column shows the ground truth damage map, reproduced from Fig. \ref{fig:fire_sar} (right panel) for convenience.
    As with the landslide, the transformer metric most accurately and confidently captures the fire extents while limiting false positives elsewhere.
    }
    \label{fig:fire_nodamage}
\end{figure*}

Fig. \ref{fig:fire_nodamage} shows disturbance metrics for the transformer, RNN, and log-ratio as in the previous section. 
As before, the transformer is the best performing, followed by the RNN. Both deep methods are significantly more accurate than the log ratio, both in terms of identifying the fire extents confidently and limiting the number of false positives elsewhere in the scene, or before the fire occurred.

Fig. \ref{fig:fire_pr} shows the PR curves for the Chilean fire scenes. 
As with the Papua New Guinea landslide, the transformer achieves the best AUC of $0.680$.
The RNN is a second with an AUC of $0.624$, and both deep methods significantly outperform the log ratio ($0.391$). 
The transformer also obtains the best $F_1$ score of $0.645$ (the best $F_1$ obtained by each method is starred). The damage maps corresponding to those best $F_1$ scores are shown in Fig. \ref{fig:fire_nodamage}, third and fourth columns. 
These binary damage maps reflect the deep learning methods' - especially the transformer's - ability to accurately map damage extents while limiting false positives.



\subsection{2024 Bangladesh Floods}\label{sec:results_flood}

\begin{figure*}[!t]
    \centering
    \begin{subfigure}{\linewidth}
        \centering
            \caption{Transformer}
        \includegraphics[width=\linewidth]{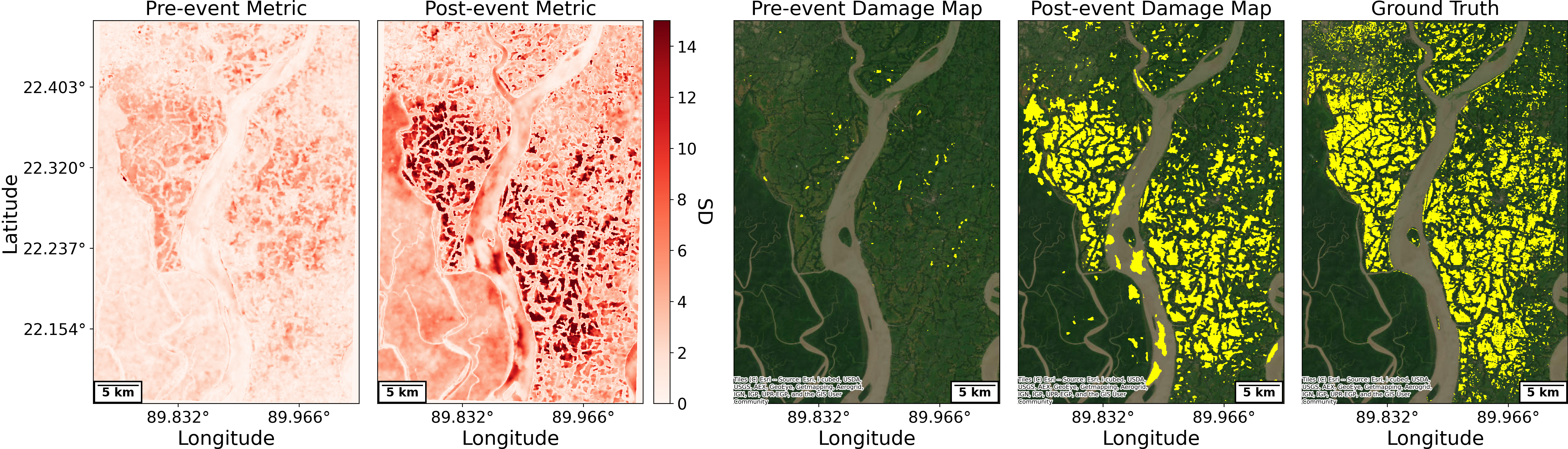}
    \end{subfigure}
    \hfill
    \begin{subfigure}{\linewidth}
        \centering
         \caption{RNN}
        \includegraphics[width=\linewidth]{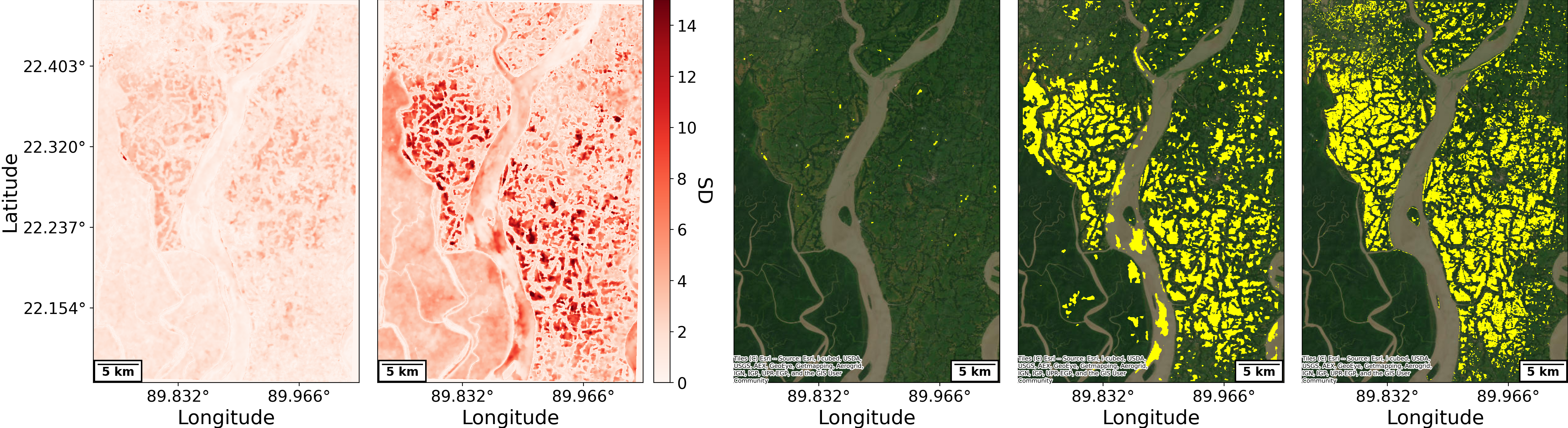}
    \end{subfigure}
    \hfill
    \begin{subfigure}{\linewidth}
        \centering
        \caption{Log Ratio}
        \includegraphics[width=\linewidth]{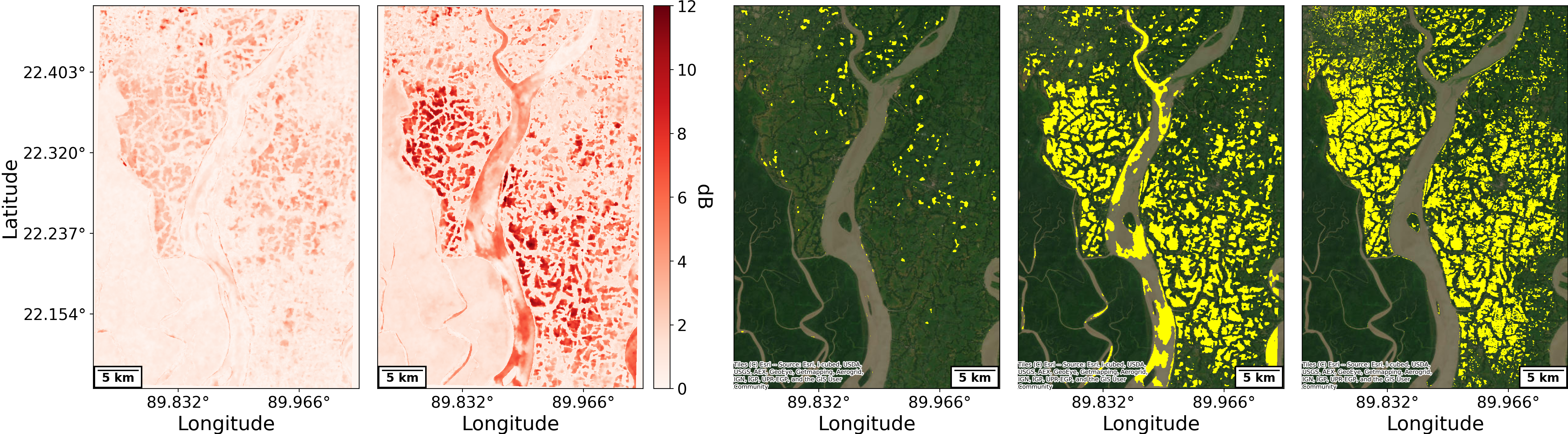}
    \end{subfigure}
    \hfill
    \caption{
    Disturbance metrics, damage maps, and ground truth data from the flooding in Bangladesh. Each row corresponds to a different model. The first two columns show the model's disturbance metric (see Section \ref{sec:metric}) before the flood (first column) and after (second column). 
    The third and fourth columns show the best (in terms of $F_1$ score, see Fig. \ref{fig:flood_pr}) binary damage maps derived from each method's metric. 
    The rightmost column shows the ground truth damage map, reproduced from Fig. \ref{fig:flood_sar} (right panel) for convenience.
    As with the other events, the transformer metric is the most performant, capturing more of the true flooding extents with fewer false positives.
    }
    \label{fig:flood_nodamage}
\end{figure*}

As in the previous two sites, we study the transformer metric, the RNN metric, and the log-ratio metric for the pre-event and post-event image in Fig. \ref{fig:flood_nodamage}. 
The disturbance is more extensive in this scene and it is difficult to draw any immediate qualitative conclusions from these maps. However, one interesting observation is that the log ratio method is somewhat confident that parts of the river are a flooded area; the deep methods do not suffer as much from this (per the ground truth damage map in Fig. \ref{fig:flood_sar}, the river is not considered a flooded area).

\begin{figure}
    \centering
    \includegraphics[width=.85\linewidth]{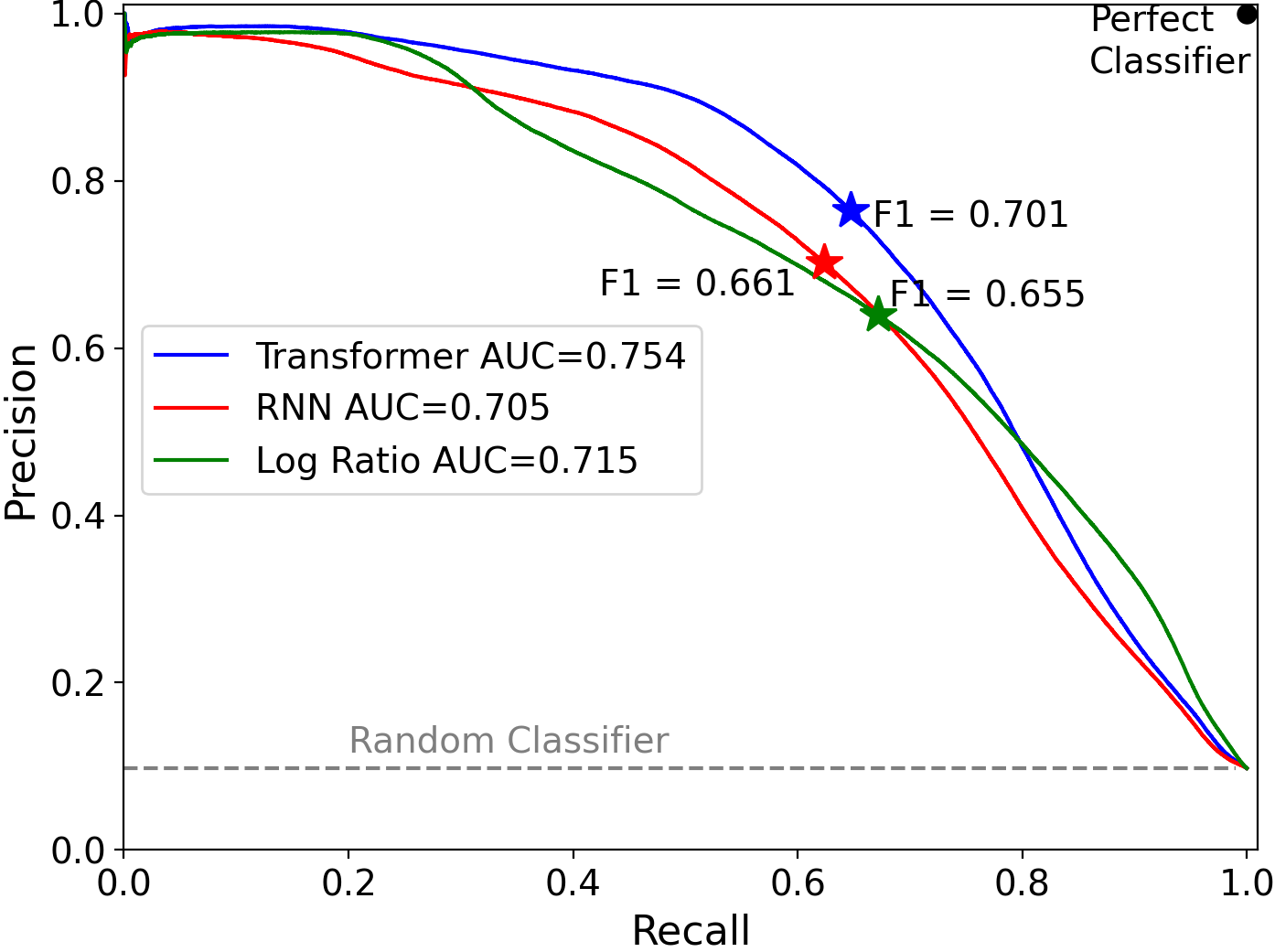}
    \caption{PR curves for the Bangladesh flood scene. 
    The transformer is the most performant model, achieving the largest area under the curve. The transformer also obtains the best $F_1$ score (starred for each model). The binary disturbance maps corresponding to the starred $F_1$ scores are shown in Fig. \ref{fig:flood_nodamage}, third and fourth columns.}
    \label{fig:flood_pr}
\end{figure}

As with the previous subsections, the quantitative results show that the transformer is the most performant of the methods. 
Fig. \ref{fig:flood_pr} shows the PR curves for the flood scene. 
Again, the transformer achieves the best AUC $0.754$.
The RNN is quite similar to the classical log ratio method (AUCs of $0.705$ and $0.715$, respectively) underscoring the improved performance from the transformer compared to the RNN. 
The transformer also obtains the best $F_1$ score ($0.701$). The disturbance delineations  corresponding to those best $F_1$ scores are shown in Fig. \ref{fig:flood_nodamage}, third and fourth columns.

\subsection{Sensitivity to $\tau$}

The preceding subsections presented PR curves (where $\tau$ is varied) and best damage maps (where $\tau$ is chosen based on $F_1$ score relative to a ground truth). However, operationally, we typically do not have access to ground truth, so choosing $\tau$ must be done \textit{a priori}. Hence, some discussion of the sensitivity of the methods to $\tau$ is warranted.

\begin{figure}[!t]
    \centering    \includegraphics[width=.85\linewidth]{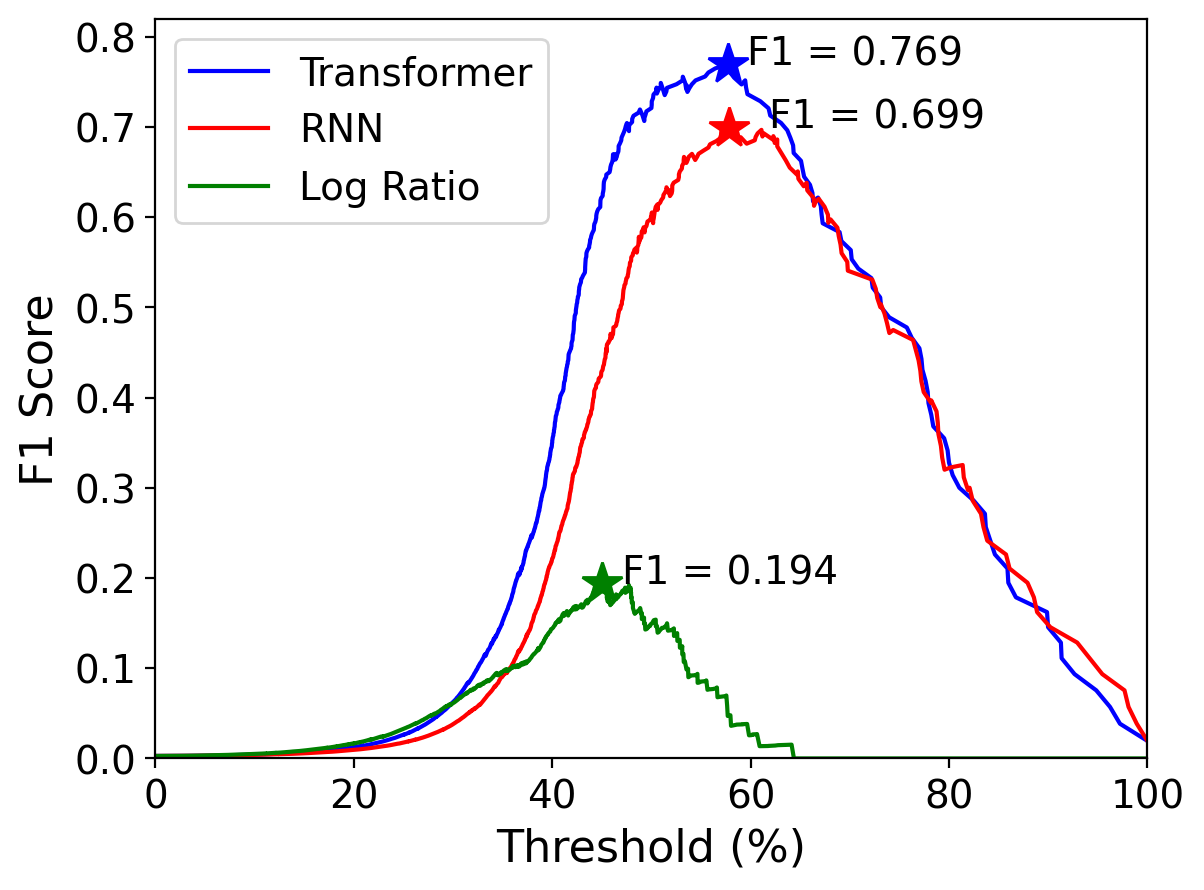}
    \caption{$F_1$ score vs. threshold for each method on the Papua New Guinea landslide. The threshold percentage is computed with respect to the maximum value (SD or dB) in the scene in order to compare all methods side-by-side. Not only does the transformer achieve the highest $F_1$ score, but also does so with a wider peak compared to the other methods, which indicates less sensitivity to the choice of threshold $\tau$.}
\label{fig:landslide_threshold}
\end{figure}

First, we evaluate the sensitivity of each method's $F_1$ score to the threshold $\tau$ for the landslide event in Fig. \ref{fig:landslide_threshold}. Natively, the deep methods' threshold is in units of standard deviations. For the log ratio, the threshold is in units of decibels. However, to compare all the methods side-by-side, we plot the thresholds on the x-axis as a percentage of the maximum threshold (where no pixels would be labeled positive). As described previously, the transformer achieves the best $F_1$ score. However, this plot illuminates that the transformer is also \textit{less sensitive to the threshold $\tau$} compared to the RNN and log ratio at their respective peaks.

\begin{figure}[!t]
    \centering
    \includegraphics[width=0.85\linewidth]{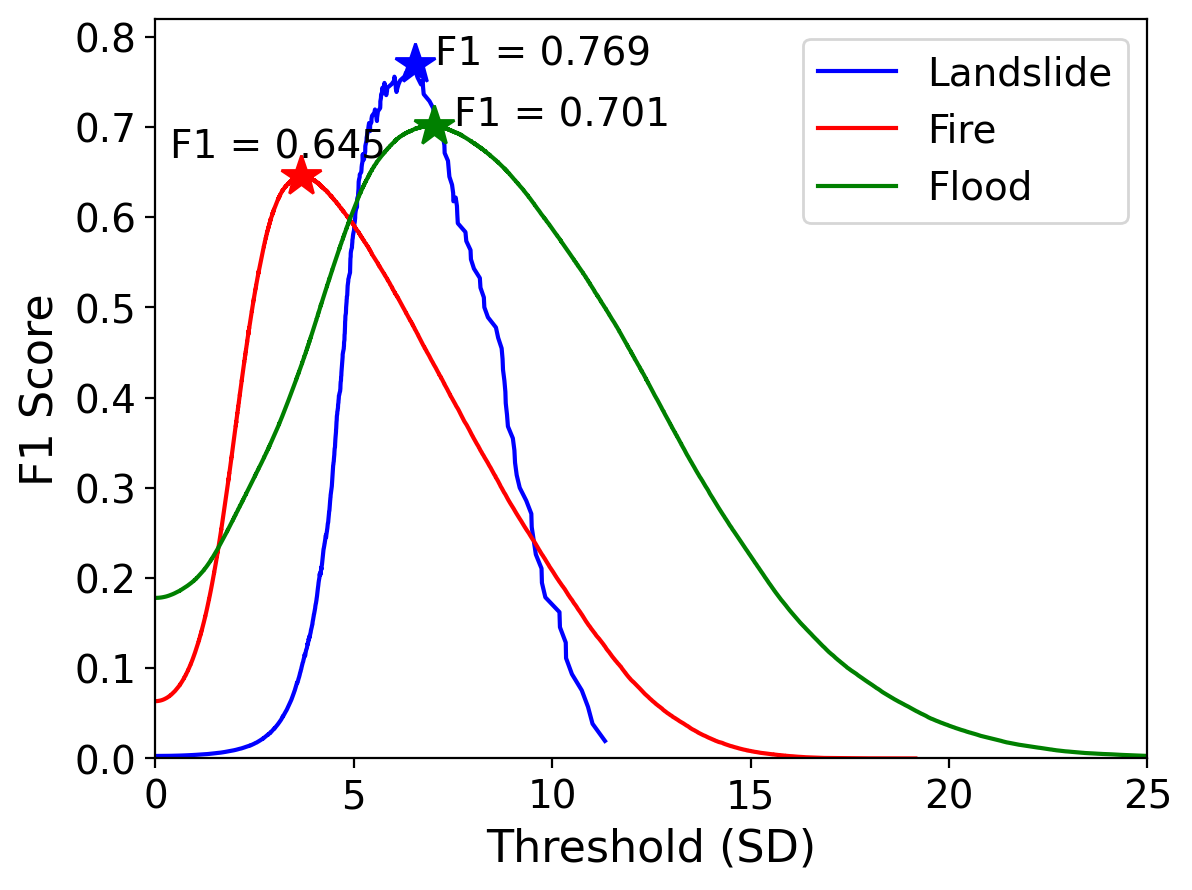}
    \caption{$F_1$ score vs thresholds for the transformer metric. Effective thresholds are consistent across all three scenes and lie in the range of $4$-$7$ standard deviations. This indicates some generality in the model and is evidence toward operational effectiveness, where the threshold $\tau$ must be chosen without knowledge of ground truth. Further study is required to corroborate this evidence.}
    \label{fig:thresholds_all}
\end{figure}

Furthermore, in Fig. \ref{fig:thresholds_all}, we show that the transformer achieves strong $F_1$ scores on all tasks at similar thresholds; a threshold of approximately $5$ standard deviations would result in $F_1$ scores around $0.6$ on all three tasks. While further validation is required, this is initial evidence that one could \emph{a priori} select $\tau \approx 5$ and expect the transformer to produce high quality disturbance delineations across environments - even when ground truth is not available. This is a critical quality for a model to possess for operational deployment. Future work will focus on further corroborating this claim on more disturbance events.

\subsection{Ablation Studies}\label{sec:ablation}

This subsection describes and empirically justifies some of the transformer and RNN hyperparameter choices made in this work. Throughout the ablation studies, we report PR AUC on each damage mapping task (landslide, fire, and flood) and bold the best result.

\subsubsection{Input and Patch Size}

Table \ref{tab:patch} compares the performance of the same transformer-based model with different input and patch sizes. The architecture and training procedure for each model in the table is the same as reported in Section \ref{sec:model_params}. 

\begingroup
\makeatletter
\long\def\@makecaption#1#2{%
\begin{center}
  {\normalfont#1}\\{\MakeUppercase{\normalfont #2}}
\end{center}
}
\makeatother

\begin{table}[!h]
    \caption{Effect of input (I) and patch sizes (P) on transformer performance}
    \centering
    \begin{tabular}{||c|c|c|c||}
        \hline
        I,P & Landslide PR AUC & Fire PR AUC & Flood PR AUC \\
        \hline
        16,8 & \textbf{0.732} & \textbf{0.680} & 0.754 \\
        \hline
        32,8 & 0.714 & 0.630 & \textbf{0.771} \\
        \hline
        32,16 &  0.630 & 0.603 & 0.744 \\
        \hline
    \end{tabular}
    \label{tab:patch}
\end{table}

\endgroup

While vision transformers for image classification tasks typically use $224 \times 224$ as the input size with a $16 \times 16$ patch size \cite{dosovitskiy2020image}, we found that a smaller input size (16) and - in particular - a smaller patch size (8) was most performant on disturbance mapping tasks. One explanation of this difference is the resolution of the data; OPERA RTC-S1 data has a resolution of 30 meters, representing a completely different scale than typical (higher resolution) image data for image classification tasks. A smaller patch size provides more spatial detail in this remote sensing setting, which has been noted in previous work \cite{cong2022satmae}. This is especially useful in disturbance mapping, where the damage extents often have a fine-grained boundary at a 30 meter resolution.

We also experiment with the input size for the RNN in Table \ref{tab:rnn_input}. 
Transformers applied to vision tasks process the input by patchifying and applying a positional encoding. 
Conversely, RNNs do not have a positional encoding. 
Since the optimal transformer in our application uses a $16 \times 16$ input size and an $8 \times 8$ patch size, we compare RNNs trained and deployed with a $16 \times 16$ and an $8 \times 8$ input size. Similar to our findings with the transformer, the smaller input size obtains better PR AUC, particularly on the landslide scene where the damaged area is smaller.

\begingroup
\makeatletter
\long\def\@makecaption#1#2{%
\begin{center}
  {\normalfont#1}\\{\MakeUppercase{\normalfont #2}}
\end{center}
}
\makeatother

\begin{table}[]
    \caption{Effect of input size on RNN performance}
    \centering
    \begin{tabular}{||c|c|c|c||}
    \hline
        Input Size & Landslide PR AUC & Fire PR AUC & Flood PR AUC \\
        \hline
        $8 \times 8$ & \textbf{0.642} &\textbf{0.624} & \textbf{0.705}  \\
        \hline
        $16 \times 16$ & 0.334 & 0.540 & 0.661 \\
        \hline
    \end{tabular}
    \label{tab:rnn_input}
\end{table}

\endgroup

\subsubsection{Model Size}

Table \ref{tab:params} compares the performance of the transformer-based model with varying architecture choices (and hence, varying number of total trainable parameters). The other modeling and training choices are the same as reported in Section \ref{sec:model_params}, except the largest model required an initial learning rate of 1e-5 instead of 1e-4 (otherwise, the loss exploded). The model with 3.3 million parameters performs the best on each disturbance mapping task, which is the model we used to report results in the previous sections. However, we expect that with significantly more training data, a larger model can improve performance over what is reported in this work. For now, we offer the comparison in Table \ref{tab:params} to justify our choices in this work, and leave scaling up the training set size and model size to future work.


\begingroup
\makeatletter
\long\def\@makecaption#1#2{%
\begin{center}
  {\normalfont#1}\\{\MakeUppercase{\normalfont #2}}
\end{center}
}
\makeatother

\begin{table}[!h]
    \caption{Effect of width of feedforward layers (FF), number of layers (L), and total number of parameters (P) on transformer performance}
    \centering
    \begin{tabular}{||c|c|c|c||}
        \hline
        FF, L, P & Landslide PRAUC & Fire PRAUC & Flood PRAUC\\
        \hline
        512, 2, 1.5M & 0.717 & 0.670 & 0.748 \\
        \hline
        768, 4, 3.3M & \textbf{0.732} & \textbf{0.680} & \textbf{0.754} \\
        \hline
        1024, 8, 7.1M & 0.728 & 0.658 & 0.744 \\
        \hline 
    \end{tabular}
    \label{tab:params}
\end{table}

\endgroup

\subsubsection{Learning Rate}

Finally, we ablate the learning rate. Table \ref{tab:lr} compares the transformer model trained at different starting learning rates (with the same learning rate decay factor of 0.1 at epoch 25). The other modeling and training choices are the same as reported in Section \ref{sec:model_params}. The starting learning rate of 1e-4 exhibits the strongest results, which corresponds to the results we report in the main text. Learning rates larger
than 1e-4 resulted in little to no training and are omitted.

\begingroup
\makeatletter
\long\def\@makecaption#1#2{%
\begin{center}
  {\normalfont#1}\\{\MakeUppercase{\normalfont #2}}
\end{center}
}
\makeatother

\begin{table}[!h]
    \caption{Effect of learning rate on transformer performance.}
    \centering
    \begin{tabular}{||c|c|c|c||}
    \hline
        LR & Landslide PR AUC & Fire PR AUC & Flood PR AUC \\
        \hline
        1e-4 & \textbf{0.732} & \textbf{0.680} & \textbf{0.754} \\
        \hline
        1e-5 & 0.725 & 0.632 & 0.746 \\
        \hline
        1e-6 & 0.648 & 0.551 & 0.678 \\
        \hline
    \end{tabular}
    \label{tab:lr}
\end{table}

\endgroup

\section{Discussion}\label{sec:discussion}

The preceding subsections demonstrate the strength of our transformer-based disturbance algorithm on three natural disasters, outperforming the RNN and log ratio on all three sites as measured by area under the precision-recall curve. 
Our results suggest that the data-driven machine learning methods, especially the transformer, are better disturbance quantification tools than the classical log-ratio approach, provided one has access to sufficient high quality training data collected for self-supervised training. Indeed, a single, relatively small transformer was able to outperform the log ratio test significantly and consistently across experiments.
Our results also illustrate the improved performance and generality of our work compared to previous deep learning-based approaches \cite{stephenson2021deep}.
Overall, our results demonstrate the effectiveness of our self-supervised transformer across different environments and disasters. 

Although we are far from verifying this approach is globally applicable, the transformer was consistently the best performing method on three different disaster events in three different parts of the world. We also note that the optimal thresholds for the transformer all fall within $4$ to $7$ standard deviations (see Fig. \ref{fig:thresholds_all}); in particular, $\tau \approx 5$ would produce $F_1 \approx 0.6$ across all three sites. This indicates some inherent generality in the model, especially for operational purposes, where the threshold $\tau$ would have to be chosen without knowledge of ground truth. In practice, the choice of threshold will depend on the application: a lower threshold will capture more true positives (that is, disturbances) at the expense of possibly more false positives, while a higher threshold will ensure fewer false positives but risk missing some true positives.

Finally, a limitation of this work is that we evaluated methods' disturbance delineation on SAR data by comparing to disturbance maps generated using optical data. There will inherently be differences and imperfections on both optical and SAR data and the delineations derived from them. Some disturbances will be less detectable (or even invisible) in one modality compared to the other. Moreover, there may be other disturbances in these scenes that are not accounted for in our ground truth damage maps, which only consider the disturbances caused by the disaster. There could be real changes in backscatter that are not related to the damage event, which would not be accounted for in our analysis. However, these limitations are difficult to negate, and due to the abruptness and scale of natural disasters, we expect the effect of non-disaster disturbances to be small. Moreover, the consistently high performance of the transformer indicates potential as a global disturbance model. 
A more complete global analysis is needed to ascertain if this transformer metric and global thresholds can be used to ascertain disturbances in a variety of environments, and be deployed for (near-)global disturbance mapping.

\section{Conclusion and Future Work}\label{sec:conclusion}

In this work, we trained deep self-supervised vision transformer on near-global OPERA RTC-S1 SAR data. 
We systematically analyzed this transformer-based metric to show it had strong agreement with externally validated data.
While many previous works on SAR disturbance require supervision or are trained on a particular location and disaster event, our model was self-supervised without needing any fine-tuning for different environments or different categories of disaster.

There are two main avenues of future work.
One is to provide systematic verification that our vision transformer approach has global applicability for the RTC-S1 data. 
This analysis includes assessing additional categories of disturbances such as logging, urbanization, and mining; analyzing the necessary baseline imagery required to delineate disturbances; and analyzing how disturbance delineations change and can be tracked over time.
The goal of such systematic analysis is to adapt this transformer disturbance algorithm for near-global disturbance monitoring through RTC-S1 inputs.

Another avenue of future work is to improve the model itself including increasing the model's depth, training the model on more data, adding more nuanced spatiotemporal features  \cite{reed2023scale, tseng2023lightweight}, a masked pretraining step \cite{he2022masked}, and increasing the window size so that the vision transformer can expand its spatial context.
The ultimate goal of this type of work is to generate a SAR foundation model
that can be fine-tuned across various wavelengths (such as C-band and L-band for the coming NISAR mission) and across multiple downstream tasks including disturbance mapping, land cover use change, classification.
Indeed, promising work has been done in this direction for multispectral data \cite{clayfoundation2024model, jakubik2023foundation, stewart2024ssl4eo} and hybrid remote sensing data \cite{tseng2023lightweight, s12_flood_methods}.
We believe the community would benefit from a SAR-specific approach, especially using the new OPERA RTC-S1 product and the forthcoming NISAR products. 

We hope this work is a step toward developing a disturbance tool that can be used to inform decision makers and emergency responders in the wake of natural disasters, and be a tool for researchers who study disturbances.

\section{Acknowledgments}
\noindent  

HHM was in part supported by the National Science Foundation Graduate Research Fellowship Program under Grant No. DGE-2034835. 
Any opinions, findings, and conclusions or recommendations expressed in this material are those of the author and do not necessarily reflect the views of the National Science Foundation. 
Part of this work supported by the 2024 UCLA JIFRESSE Summer Internship Program (JSIP). 
Part of this research was carried out at the Jet Propulsion Laboratory, California Institute of Technology, under a contract with the National Aeronautics and Space Administration (80NM0018D0004). 
Copyright 2024 by the California Institute of Technology. ALL RIGHTS RESERVED. United States Government Sponsorship acknowledged. Any commercial use must be negotiated with the Office of Technology Transfer at the California Institute of Technology.
The OPERA project, managed by the Jet Propulsion Laboratory and funded by the Satellite Needs Working Group, creates remote sensing products to address Earth observation needs across U.S. civilian federal agencies.

We appreciate early feedback and suggestions about this work from the OPERA Project team, specifically David Bekaert, Steven Chan, Richard West, Talib Oliver-Cabrera, and Jungkyo Jung.
OPERA is funded by NASA under the Satellite Needs Working group cycle-2 activities. This work contains modified Copernicus Sentinel-1 data, processed by ESA and OPERA. PlanetScope images were provided by Planet's Education and Research Program to ALH.
\bibliography{references}
\bibliographystyle{IEEEtran}

\newpage

\begin{IEEEbiography}[{\includegraphics[width=1in,height =1.25in,clip]{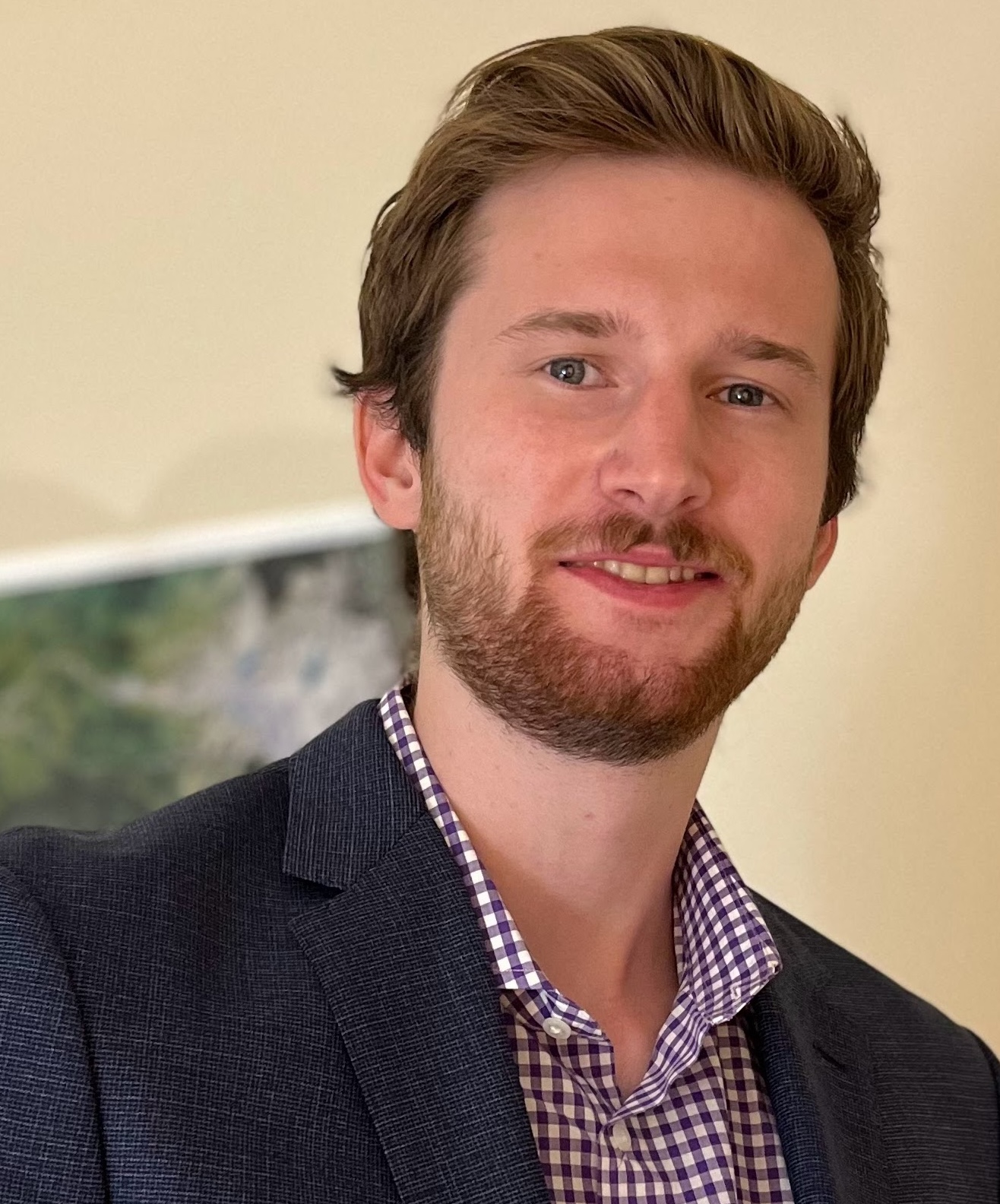}}]{Harris Hardiman-Mostow}
is a mathematics Ph.D. candidate at UCLA. He is a recipient of a National Science Foundation Graduate Research Fellowship. He received a B.S. \textit{summa cum laude} with a double major in mathematics and mechanical engineering from Tufts University in 2021, and an M.A. in mathematics from UCLA in 2023. His research focuses on developing novel deep learning and graph-based learning algorithms and applying them to problems in sensing, the environment, and climate.
\end{IEEEbiography}

\vspace{-13cm}

\begin{IEEEbiography}[{\includegraphics[width=1in,height =1.25in,clip]{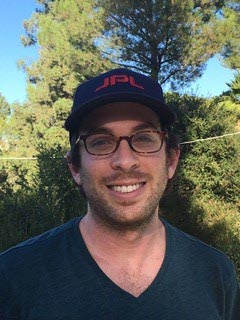}}]{Charles Marshak}
is a Signal Analysis engineer at Jet Propulsion Laboratory within Radar Science and Engineering.
He supports The Advanced Rapid Imaging and Analysis (ARIA) and Observational Products for End-Users from Remote Sensing Analysis (OPERA) projects.
He earned his Ph.D. at UCLA under the supervision of Andrea Bertozzi and Mason Porter in applied and computational mathematics.
His research interests lie at the intersection of machine learning and SAR remote sensing.
\end{IEEEbiography}

\vspace{-13cm}

\begin{IEEEbiography}
[{\includegraphics[width=1in,height =1.25in,clip]{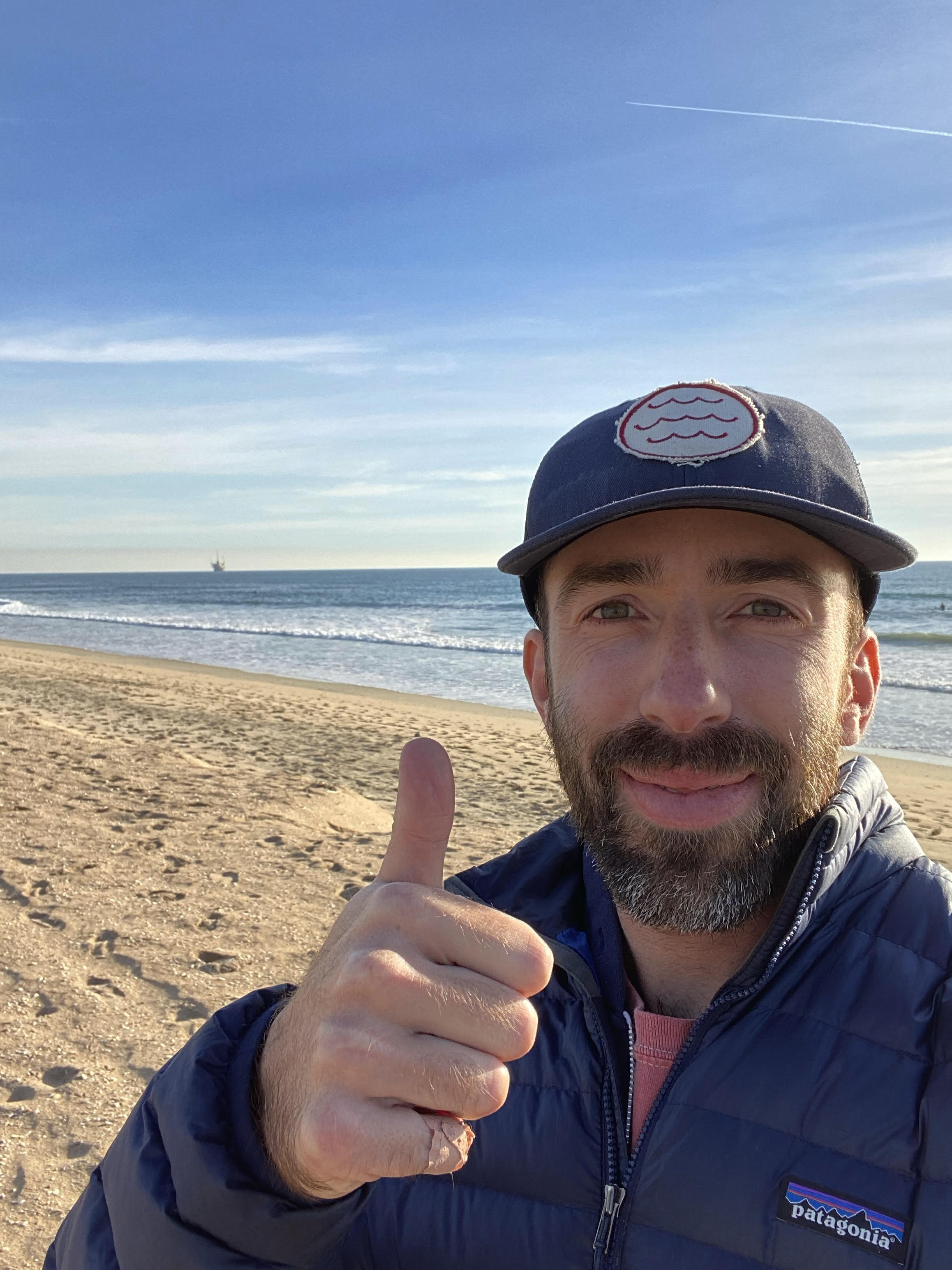}}]{Alexander L. Handwerger} received the B.A. degree in Earth Sciences from Boston University, Boston, MA, USA, in 2008, and the Ph.D. degree in Geological Sciences from the University of Oregon, Eugene, OR, USA, in 2015. He is currently a Research Scientist at the Jet Propulsion Laboratory, California Institute of Technology and an Assistant Researcher at the Joint Institute for Regional Earth System Science and Engineering (JIFRESSE), University of California, Los Angeles, CA, USA. He is the Deputy Project Scientist for the OPERA project. His research interests include landslides, rock glaciers, and satellite and airborne interferometric synthetic aperture radar.

\end{IEEEbiography}

\end{document}